\documentclass{article}

\PassOptionsToPackage{numbers, compress}{natbib}

\usepackage[preprint]{neurips_2020}

\usepackage[utf8]{inputenc} 
\usepackage[T1]{fontenc}    
\usepackage{hyperref}       
\usepackage{url}            
\usepackage{booktabs}       
\usepackage{amsfonts}       
\usepackage{nicefrac}       
\usepackage{microtype}      
\usepackage{mathtools}

\usepackage{natbib}

\usepackage{breakurl}
\usepackage[font=small]{caption}
\usepackage{wrapfig}
\usepackage{textcomp}
\usepackage{blindtext}

\usepackage{multirow}
\usepackage{dsfont}
\usepackage{multicol}
\usepackage{amsmath}
\usepackage{amsthm}
\usepackage{amssymb}
\usepackage{bm}
\usepackage{mathrsfs}
\usepackage{xcolor}
\usepackage{xspace}
\usepackage{colortbl}
\usepackage[labelformat=simple]{subcaption}
\usepackage[hang,flushmargin]{footmisc}

\usepackage{color}
\usepackage{mathtools}
\usepackage{soul}

\usepackage[ruled,noline,noend]{algorithm2e}
\SetNlSty{}{}{}
\let\oldnl\nl
\newcommand\nonl{%
  \renewcommand{\nl}{\let\nl\oldnl}}

\usepackage{pgfplots}
\pgfplotsset{
  compat=newest,
  plot coordinates/math parser=false,
  tick label style={font=\footnotesize, /pgf/number format/fixed},
  label style={font=\small},
  legend style={font=\small},
  every axis/.append style={
    tick align=outside,
    clip mode=individual,
    scaled ticks=false,
    thick,
    tick style={semithick, black}
  }
}

\pgfkeys{/pgf/number format/.cd, set thousands separator={\,}}

\newlength\figurewidth
\newlength\figureheight

\newlength\sbsfigurewidth
\newlength\sbsfigureheight

\newcommand{\LOVE}{\acro{LOVE}\xspace}
\newcommand{\QR}{\acro{QR}\xspace}
\newcommand{\bo}{\acro{BO}\xspace}

\newcommand{\mdp}{\acro{MDP}\xspace}

\newcommand{\gp}{\acro{GP}\xspace}
\newcommand{\gps}{\acro{GP}s\xspace}

\newcommand{\ei}{\acro{EI}\xspace}

\newcommand{\kg}{\acro{kg}\xspace}

\newcommand{\ens}{\acro{ENS}\xspace}
\newcommand{\eno}{\acro{ENO}\xspace}
\newcommand{\af}{\acro{AF}\xspace}

\newcommand{\qei}{$q$-\acro{EI}\xspace}
\newcommand{\ets}{\acro{ETS}\xspace}
\newcommand{\gh}{\acro{GH}\xspace}
\newcommand{\gap}{\acro{GAP}\xspace}

\newcommand{\lda}{\acro{LDA}\xspace}
\newcommand{\svm}{\acro{SVM}\xspace}

\newcommand{\binoc}{\acro{BINOCULARS}\xspace}
\newcommand{\glasses}{\acro{GLASSES}\xspace}
\newcommand{\botorch}{\texttt{BoTorch}\xspace}
\newcommand{\gpytorch}{\texttt{GPyTorch}\xspace}
\DeclareMathOperator*{\argmax}{arg\,max}
\newcommand{\bpi}{{\boldsymbol{\pi}}}

\newcommand{\R}{\mathbb{R}}
\newcommand{\mc}[1]{\mathcal{#1}}
\newcommand{\data}{\mc{D}}

\newcommand{\given}{\, | \,}
\newcommand{\E}{\mathbb{E}}

\newcommand{\acro}[1]{\textsc{\MakeLowercase{#1}}}

\newif\ifboldmatrix
\boldmatrixfalse
\ifboldmatrix\newcommand{\boldmatrix}[1]{\mathbf{#1}}\else\newcommand{\boldmatrix}[1]{#1}\fi
\newcommand{\dset}{\ensuremath{\mathcal D}}
\newcommand{\bk}{\ensuremath{\mathbf{k}}}
\newcommand{\bv}{\ensuremath{\mathbf{v}}}
\newcommand{\X}{\ensuremath{\boldmatrix{X}}}
\newcommand{\bigo}[1]{\mathcal O ( #1 )}

\theoremstyle{definition}

\theoremstyle{plain}

\newtheorem{proposition}{Proposition}

\usepackage{enumitem}
\usepackage{tikz}
\usetikzlibrary{positioning}

\newcommand{\papertitle}{Efficient Nonmyopic Bayesian Optimization via One-Shot Multi-Step Trees}
\title{\papertitle}

\author{%
  Shali Jiang\thanks{Equal contribution.} \\
  Washington University\\
  \texttt{jiang.s@wustl.edu} 
  \And
  Daniel R. Jiang$^*$ \\
  Facebook \\
  \texttt{drjiang@fb.com} 
  \And
  Maximilian Balandat$^*$ \\
    Facebook \\
    \texttt{balandat@fb.com}
  \And
  Brian Karrer \\
    Facebook \\
   \texttt{briankarrer@fb.com} 
   \And
   Jacob R. Gardner \\
   University of Pennsylvania \\
   \texttt{gardner.jake@gmail.com} 
   \And
   Roman Garnett \\
   Washington University\\
   \texttt{garnett@wustl.edu} 
}

\begin{document}

\maketitle

\begin{abstract}
Bayesian optimization is a sequential decision making framework for optimizing expensive-to-evaluate black-box functions. Computing a full lookahead policy amounts to solving a highly intractable stochastic dynamic program. Myopic approaches, such as expected improvement, are often adopted in practice, but they ignore the long-term impact of the immediate decision. Existing nonmyopic approaches are mostly heuristic and/or computationally expensive. In this paper, we provide the first efficient implementation of general multi-step lookahead Bayesian optimization, formulated as a sequence of nested optimization problems within a multi-step scenario tree. Instead of solving these problems in a nested way, we equivalently optimize all decision variables in the full tree jointly, in a ``one-shot'' fashion. Combining this with an efficient method for implementing multi-step Gaussian process ``fantasization,'' we demonstrate that multi-step expected improvement is computationally tractable and exhibits performance superior to existing methods on a wide range of benchmarks.
\end{abstract}

\section{Introduction}
\label{section:intro}
Bayesian optimization (\bo) is a powerful technique for optimizing expensive-to-evaluate black-box functions.  
Important applications include materials design \citep{zhang2020bayesian}, drug discovery \citep{griffiths2020constrained}, machine learning hyperparameter tuning \citep{snoek2012practical}, neural architecture search \citep{kandasamy2018neural, zhang2019d}, etc. 
\bo operates by constructing a \emph{surrogate model} for the target function, typically a Gaussian process (\gp), and then using a 
cheap-to-evaluate \emph{acquisition function} (\af) to guide iterative queries of the target function until a predefined budget is expended.
We refer to \cite{shahriari2016taking} for a literature survey. 

Most of the existing acquisition policies are only one-step optimal, that is, optimal if the decision horizon were one. An example is \emph{expected improvement} (\ei) \citep{movckus1974bayesian}, arguably the most widely used policy. Such \emph{myopic} policies only consider the immediate utility of the decision, ignoring the long-term impact of exploration. Despite the sub-optimal balancing of exploration and exploitation, they are widely used in practice due to their simplicity and computational efficiency.

When the query budget is explicitly considered, \bo can be formulated as a Markov decision process (\mdp), whose optimal policy maximizes the expected utility of the terminal dataset \citep{lam2016bayesian, jiang2019binoculars}.
However, solving the \mdp is generally intractable due to the set-based, uncountable state space, uncountable action space, and potentially long decision horizon. There has been recent interest in developing nonmyopic policies \citep{gonzalez2016glasses, lam2016bayesian, wu2019practical, yue2019why}, but these policies are often heuristic in nature or computationally expensive. A recent work known as \binoc \citep{jiang2019binoculars} achieved both efficiency and a certain degree of nonmyopia by maximizing a lower bound of the multi-step expected utility. 
However, a general implementation of multi-step lookahead for \bo has, to our knowledge, not been attempted before. 

\textbf{Main Contributions.} Our work makes progress on the intractable multi-step formulation of \bo through the following methodological and empirical contributions:
\begin{itemize}[leftmargin=*]
    \item \textit{One-shot multi-step trees.} We introduce a novel, scenario tree-based acquisition function for \bo that performs an approximate, multi-step lookahead. Leveraging the reparameterization trick, we propose a way to jointly optimize all decision variables in the multi-step tree in a \emph{one-shot} fashion, without resorting to explicit dynamic programming recursions involving nested expectations and maximizations. Our tree formulation is fully differentiable, and we compute gradients using auto-differentiation, permitting the use of gradient-based optimization.
    \item \textit{Fast-fantasies and parallelism.} 
    Our multi-step scenario tree is built by recursively sampling from the \gp posterior and conditioning on the sampled function values (``fantasies''). 
    This tree grows exponentially in size with the number of lookahead steps. 
    While our algorithm cannot negate this reality, our novel efficient linear algebra methods for conditioning the \gp model combined with a highly parallelizable implementation on accelerated hardware allows us to tackle the problem at practical wall times for moderate lookahead horizons (less than 5).
    \item \textit{Improved optimization performance.} Using our method, we are able to achieve significant improvements in optimization performance over one-step \ei and \binoc on a range of benchmarks, while maintaining competitive wall times. To further improve scalability, we study two special cases of our general framework which are of linear growth in the lookahead horizon. We empirically show that these alternatives perform surprisingly well in practice.
\end{itemize}
We set up our problem setting in Section \ref{sec:bayesoptimalpolicy} and propose the one-shot multi-step trees in Section \ref{sec:OST}. We discuss how we achieve fast evaluation and optimization of these trees in Section \ref{sec:fastdifffantasies}. In Section \ref{sec:pseudo}, we show some notable instances of multi-step trees and make connections to related work in Section \ref{section:related}. We present our empirical results in Section \ref{sec:exper} and conclude in Section \ref{section:conclusion}.

\section{Bayesian Optimal Policy}
\label{sec:bayesoptimalpolicy}

We consider an optimization problem 
\begin{equation}
    x^* \in \textstyle{ \argmax_{x\in \mc{X}} } f(x), 
    \label{eq:mainobj}
\end{equation}
where $\mc{X} \subset \R^d$ and $f(x)$ is an expensive-to-evaluate black-box function. 
Suppose we have collected a (possibly empty) set of initial observations $\data_0$ and a probabilistic surrogate model of $f$ that provides a joint distribution over outcomes $p(Y \,|\, X, \data_0)$ for all finite subsets of the design space $X \subset \mathcal X$.
We need to reason about the locations to query the function next in order to find the maximum, given the knowledge of the remaining budget.
Suppose that the location with maximum observed function value is returned at the end of the \emph{decision horizon}~$k$. A natural utility function for sequentially solving (\ref{eq:mainobj}) is
\begin{equation}
    u(\data_k) = \textstyle{ \max_{(x,y)\in\data_k} } y,  \label{eq:utility_definition}
\end{equation}
where $\data_k$ is the sequence of observations up to step~$k$, defined recursively by $\data_i = \data_{i-1} \cup \{(x_i, y_i)\}$ for $i = 1, 2, \ldots, k$. Due to uncertainties in the future unobserved events, $\data_1, \data_2, \ldots, \data_k$ are random quantities. A policy $\boldsymbol{\pi} = (\pi_1, \pi_2, \ldots, \pi_{k})$ is a collection of decision functions, where at period $i$, the function $\pi_i$ maps the dataset $\data_{i-1}$ to the query point $x_i$. Our objective function is $\sup_\bpi \E [  u(\data_k^\bpi) ]$, where $\{\data_i^\bpi\}$ is the sequence of datasets generated when following $\bpi$.

For any dataset $\data$ and query point $x\in\mathcal X$, define the one-step marginal value as
\begin{align}
    v_1(x \given \data) = \E_y\bigl[u(\data \cup \{(x, y)\}) - u(\data) \given x, \data\bigr].  \label{eq:marginal_value}
\end{align}
Note that under the utility definition \eqref{eq:utility_definition}, $v_1(x \,|\, \data)$
is precisely the expected improvement (\ei) acquisition function \citep{movckus1975bayesian}. It is well-known that the $k$-step problem can be decomposed via the Bellman recursion \citep{lam2016bayesian, jiang2019binoculars}:
\begin{align}
        v_t(x \given \data) 
        =
        v_1(x \given \data) 
        + 
        \E_y[ \textstyle{ \max_{x'} } v_{t-1}(x' \given \data \cup \{(x, y)\} ) ], \label{eq:bellman_equation}
    \end{align}
for $t=2,3, \ldots, k$. Our $k$-step lookahead acquisition function is $v_k(x\given \data)$, meaning that a maximizer in $\argmax_{x} v_k(x\given \data)$ is the recommended next point to query.

If we are allowed to evaluate multiple points $X=\{x^{(1)}, \dots, x^{(q)}\}$ in each iteration, we replace $v$ with a batch value function $V$. 
For $k=1$ and batch size $|X| = q$, we have
    \[
        V_1^q(X\given \data ) = \E_{y^{(1)},\ldots,y^{(q)}} \bigl[u(\data \cup \{(x^{(1)}, y^{(1)}), \ldots, (x^{(q)}, y^{(q)}) \}) - u(\data)\given X, \data \bigr],
    \]
    which is known as $q$-\ei in the literature \citep{ginsbourger2010kriging, wang2016parallel}.
For general $k$, $V_k$ is the exact analogue of 
\eqref{eq:bellman_equation}; we capitalize $v$ and $x$ to indicate expected value of a batch of points. 
While we only consider the fully adaptive setting ($q=1$) in this paper, we will make use of the batch policy for approximation. 

\section{One-Shot Optimization of Multi-Step Trees}
\label{sec:OST}
In this section, we describe our multi-step lookahead acquisition function, a differentiable, tree-based approximation to $v_k(x\given\data)$. We then propose a one-shot optimization technique for effectively optimizing the acquisition function and extracting a first-stage decision.
\subsection{Multi-Step Trees}
\label{subsec:OST:MST}

Solving the $k$-step problem requires recursive maximization and integration over continuous domains:
\begin{equation}
    v_k(x\given \data) 
    = 
    v_1(x \given \data) 
    +
    \E_y\Big[\max_{x_2} \Big\{v_1(x_2 \given \data_1) + 
    \E_{y_2}\big[ \max_{x_3} \big\{ v_1(x_3 \given \data_2) +
    \cdots  \Big].
    \label{eq:recursive_max_and_exp}
\end{equation}
Since under a \gp surrogate, these nested expectations are analytically intractable (except the last step for  \ei), we must resort to numerical integration. 
If we use Monte Carlo integration, this essentially means building a discrete scenario tree (Figure \ref{fig:decision_tree}), where each branch in a node corresponds to a particular \emph{fantasized} outcome drawn from the model posterior, and then averaging across scenarios. Letting $m_t$, $t=1,\dots, k-1$ denote the number of \emph{fantasy samples} from the posterior in step $t$, 
we have the approximation 
\[
    \bar{v}_k(x\given \data) 
    = 
    v_1(x \given \data) 
    +
    \frac{1}{m_1}\sum_{j_1=1}^{m_1} \Big[\max_{x_2} \Big\{v_1(x_2 \given \data_1^{j_1}) + 
    \frac{1}{m_2}\sum_{j_2=1}^{m_2} \big[ \max_{x_3} \big\{ v_1(x_3 \given \data_2^{j_1 j_2}) +
    \cdots  \Big]. 
\label{eq:multistagesaa}
\]
where $\data^{j_1}_1 = \data \cup \{(x, y^{j_1}) \}$, 
$\data^{j_1\dotsc j_t}_t = \data^{j_1\dotsc j_{t-1}}_{t-1} \cup \{(x_{t}^{j_1\dotsc j_{t-1}}, y_t^{j_1\dotsc j_{t}})\}$, with fantasy samples $y_t^{j_1\dotsc j_{t}} \sim p(y_t \given x_{t}^{j_1\dotsc j_t}, \data^{j_1\dotsc j_{t-1}}_{t-1})$. 
As the distribution of the fantasy samples depends on the query locations $x, x_1, x_2, \ldots$, we cannot directly optimize $\bar{v}_k(x\given\data)$. To make $\bar{v}_k(x\given\data)$ amenable to optimization, we leverage the re-parameterization trick \cite{kingma2013reparam, wilson2018maxbo} to write $y = h_{\data}(x, z)$, where $h_{\data}$ is a deterministic function and $z$ is a random variable independent of both $x$ and $\mathcal{D}$. Specifically, for a \gp posterior, we have $h_{\data}(x, z) = \mu_{\data}(x) + L_{\data}(x) z$,
%
%
where $\mu_{\data}(x)$ is the posterior mean, $L_\data(x)$ is a root decomposition of the posterior covariance $\Sigma_\data(x)$ such that $L_\data(x)L_\data^T(x) = \Sigma_\data(x)$, and $z \sim \mathcal{N}(0, I)$. 
Since a \gp conditioned on additional samples remains a \gp, we can perform a similar re-parameterization for every dataset $\data^{j_1\dotsc j_{t}}_t$ in the tree. 
We refer to the $z$'s as \emph{base samples}.


\tikzset{
  decision/.style = {minimum size=1.5em,shape=rectangle,draw,yshift=-0.5em},
  empty/.style = {draw=none}
}
\begin{figure}
    \centering
\begin{tikzpicture}[
      grow                    = right,
      sibling distance        = 2.1em,
      level distance          = 5.5em,
    ]
    \node [decision] {$\data$}
    child { node [empty] {$x$} 
        child { node [empty] {$z_1^3$} 
            child { node [empty] {$\cdots$ } 
            }
        }
        child { node [empty] {$z_1^2$} 
            child { node [empty] {$x^2_2$} 
                child { node [empty] {$z_2^{2,3}$} 
                    child { node [empty] {$x^{2,3}_3$} 
                        child { node [empty] {$\dots$} }
                        child { node [empty] {$\cdots$} }
                        child { node [empty] {$\cdots$} }
                    }
                }
                child { node [empty] {$z_2^{2,2}$} 
                    child { node [empty] {$x^{2,2}_3$} 
                        child { node [empty] {$\cdots$} }
                        child { node [empty] {$\cdots$} }
                        child { node [empty] {$\cdots$} }
                    }
                }
                child { node [empty] {$z_2^{2,1}$} 
                    child { node [empty] {$x^{2,1}_3$} 
                        child { node [empty] {$\cdots$} }
                        child { node [empty] {$\cdots$} }
                        child { node [empty] {$\dots$} }
                    }
                }
            }
        }
        child { node [empty] {$z_1^1$} 
            child { node [empty] {$\cdots$} 
            }
        }
    };
  \end{tikzpicture} 
      \caption{Illustration of the decision tree with three base samples in each stage.} 
    \label{fig:decision_tree}
\end{figure}
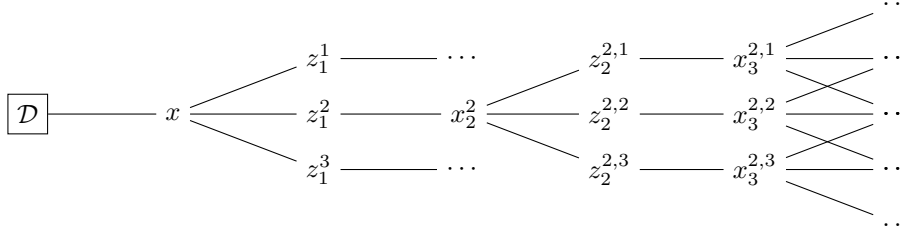
%

\subsection{One-Shot Optimization}
\label{subsec:OST:OSO}
Despite re-parameterizing $\bar{v}_k(x\given\data)$ using base samples, it still involves nested maximization steps. Particularly when each optimization must be performed numerically using sequential approaches (as is the case when auto-differentiation and gradient-based methods are used), this becomes cumbersome. Observe that conditional on the base samples, $\bar{v}_k$ is a \emph{deterministic} function of the decision variables. 
%
%
\begin{proposition}
\label{prop:oneshot}
Fix a set of base samples and consider $\bar{v}_k(x\given\data)$. Let $x_t^{j_1\dots j_{t-1}}$ be an instance of $x_t$ for each realization of $\data_{t-1}^{j_1\dots j_{t-1}}$ and let
\begin{multline}
    x^*, \mathbf{x}^*_2, \mathbf{x}^*_3, \dots, \mathbf{x}^*_k
    =
    \argmax_{x, \mathbf{x}_2, \mathbf{x}_3, \dots, \mathbf{x}_k} \Biggl\{
    v_1(x \given \data) 
    +
   \frac{1}{m_1}\sum_{j_1=1}^{m_1} v_1(x_2^{j_1} \given \data_1^{j_1}) 
   + \cdots  
   +  \\
    \frac{1}{\prod_{\ell=1}^{k-1} m_{\ell}}
    \sum_{j_1=1}^{m_1} 
    \cdots 
    \sum_{j_{k-1}=1}^{m_{k-1}}  
    v_1(x_k^{j_1\cdots j_{k-1}} \given \data_{k-1}^{j_1\cdots j_{k-1}})  \Biggr\}, \label{eq:one-shot-objective}
\end{multline}
where we compactly represent $\mathbf{x}_2 = \{x^{j_1}_2 \}_{j_1=1\dots m_1}$, 
$\mathbf{x}_3 = \{x^{j_1 j_2}_3\}_{j_1=1\dots m_1, j_2=1\dots m_2}$, and so on.
Then, 
$x^* = \argmax_x \bar{v}_k(x \given \data)$.
\end{proposition}
Proposition~\ref{prop:oneshot} suggests that rather than solving a nested optimization problem, we can solve a joint optimization problem of higher dimension and subsequently extract the optimizer. We call this the \emph{one-shot multi-step} approach. A single-stage version of this was used in \cite{balandat2019botorch} for optimizing the Knowledge Gradient (\kg) acquisition function \citep{wu2016parallel}, which also has a nested maximization (of the posterior mean). Here, we generalize the idea to its full extent for efficient multi-step \bo.

\section{Fast, Differentiable, Multi-Step Fantasization}
\label{sec:fastdifffantasies}
Computing the one-shot objective~\eqref{eq:one-shot-objective}
requires us to repeatedly condition the model on the fantasy samples as we traverse the tree to deeper levels. Our ability to solve multi-step lookahead problems efficiently is made feasible by linear algebra insights and careful use of efficient batched computation on modern parallelizable hardware.
Typically, conditioning a \gp on additional data in a computationally efficient fashion is done by performing rank-1 updates to the Cholesky decomposition of the input covariance matrix. 
In this paper, we develop a related approach, which we call \emph{multi-step fast fantasies}, in order to efficiently construct fantasy models for \gpytorch~\citep{gardner2018gpytorch} \gp models representing the full lookahead tree. A core ingredient of this approach is a novel linear algebra method for efficiently updating \gpytorch's \LOVE caches~\citep{pleiss2018love} for posterior inference in each step. 

\subsection{Background: Lanczos Variance Estimates}
\label{subsec:FastFantasies:Background}

We start by providing a brief review of the main concepts for the Lanczos Variance Estimates (\LOVE) as introduced in~\cite{pleiss2018constant}. The \gp predictive covariance between two test points $x^*_i$ and $x^*_j$ is given by:
\[
  k_{f\mid\dset}(x^*_i, x^*_j) = k_{x^*_i x^*_j} - \bk_{\X \! x^*_i}^\top (K_{\X\!\X} + \Sigma)^{-1} \bk_{\X \! x^*_j},
\]
$X = (x_1, \dotsc, x_n)$ is the set of training points, $K_{XX}$ is the kernel matrix at~$X$, and $\Sigma$ is the noise covariance.\footnote{Typically, $\Sigma = \sigma^2 I$, but other formulations, including heteroskedastic noise models, are also compatible with fast fantasies described here.} 
\LOVE achieves fast (co-)variances by decomposing $K_{\X\!\X} + \Sigma = RR^{\top}$ in $\bigo{r\nu(K_{\X\!\X})}$ time, where $R \in \mathbb{R}^{n\times r}$ and  $\nu(K_{\X\!\X})$ is the time complexity of a matrix vector multiplication $K_{\X\!\X}v$. This allows us to compute the second term of the predictive covariance as:
\[
  k_{f\mid\dset}(x^*_i, x^*_j) = k_{x^*_i x^*_j} - \bk_{\X \! x^*_i}^\top R^{-\top}R^{-1} \bk_{\X \! x^*_j},
\]
where $R^{-1}$ denotes a pseudoinverse if~$R$ is low-rank.\footnote{Additional approximations can be performed when using Spectral Kernel Interpolation (SKI), which result in constant time predictive covariances. For simplicity, we only detail the case of exact \gps here.} 
The main operation to perform is decomposing $\tilde{K}_{\X\!\X} = RR^{\top}$, where $\tilde{K}_{\X\X} := K_{\X\!\X} + \Sigma \in \mathbb{R}^{n \times n}$. 
Computing this decomposition can be done from scratch in $\bigo{nr^{2}}$ time. After forming~$R$, additional $\bigo{nr^{2}}$ time is required to perform a \QR decomposition of~$R$ so that a linear least squares systems can be solved efficiently (i.e., approximate $R^{-1}$). $R$ and its \QR decomposition are referred to as the \emph{\LOVE cache}.

\subsection{Fast Cache Updates}
\label{subsec:FastFantasies:FastCache}

If $R$ were a full Cholesky decomposition of $\tilde{K}_{\X\X}$, it could be updated in $\bigo{n^2}$ time using well-known procedures for rank 1 updates to Cholesky decompositions. This is advantageous, because computing the Cholesky decomposition requires $\bigo{n^3}$ time. However, for dense matrices, the \LOVE cache requires only $\bigo{n^2r}$ time to compute. Therefore, for it to be useful to multi-step lookahead, we must demonstrate that it can be updated in $o(n^2)$ time.
Updating the \LOVE caches is in particular complicated by the fact that $R$ is not necessarily triangular (or even square). Therefore, unlike with a Cholesky decomposition, updating $R$ itself in linear time is not sufficient, as recomputing a \QR decomposition of $R$ to update the pseudoinverse $R^{\dagger}$ would itself take quadratic time. In the Appendix \ref{appdx:sec:FastFantasies}, we demonstrate that the following proposition is true:
\begin{proposition}
Suppose $(K_{\X\!\X} + \Sigma)^{-1}$ has been decomposed using LOVE into $R^{-\top}R^{-1}$, with $R^{-1} \in \mathbb{R}^{n \times r}$. Suppose we wish to augment $\X$ with $q$ data points, thereby augmenting $K_{\X\!\X}$ with $q$ rows and columns, yielding $K_{\hat{\X}\!\hat{\X}}$. A rank $r+q$ decomposition $\hat{R}^{-1}$ of the inverse, $\hat{R}^{-\top}\hat{R}^{-1} \approx (K_{\hat{\X}\!\hat{\X}} + \Sigma)^{-1}$, can be computed from $R$ in $\bigo{nrq}$ time.
\end{proposition}

\begin{figure*}[t]
    \begin{minipage}[c]{0.45\columnwidth}
        \centering
        \includegraphics[width=\textwidth]{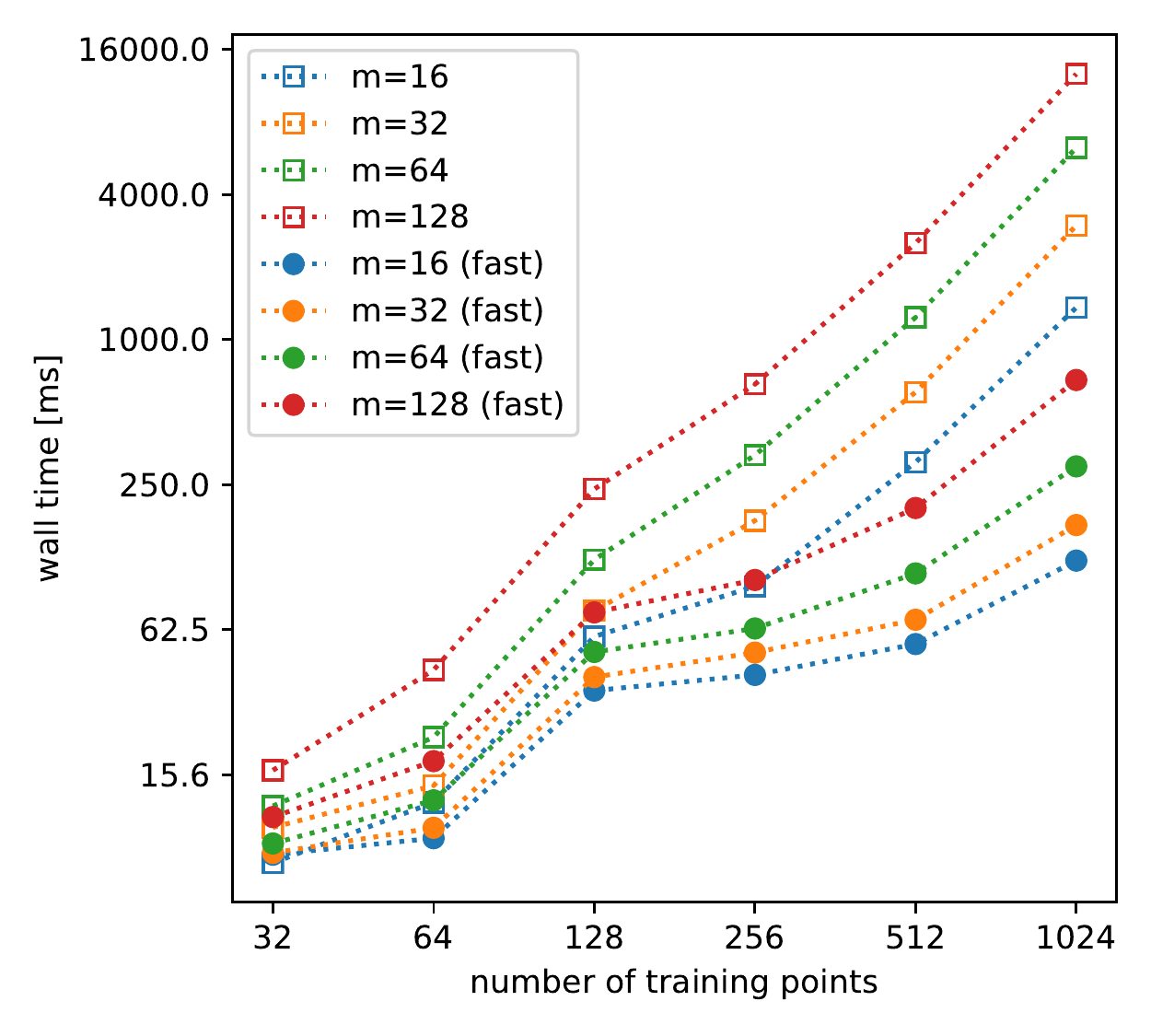}\\[-1.5ex]
        \caption{\acro{CPU} times for constructing fantasy model and evaluating its posterior at a single point (variance negligible relative to the mean). 
        }
        \label{fig:ParallelHardware:FastFantasies:WallTimes2}
    \end{minipage}
    \hfill
    \begin{minipage}[c]{0.528\columnwidth}
        \begin{algorithm}[H]
        \small
        \SetKwFunction{step}{\textsc{Step}}
        \SetKwFunction{sample}{\textsc{sample}}
        \SetKwFunction{stageval}{\textsc{value}}
        \SetKwFunction{correlate}{\textsc{correlate}}
        \SetKwFunction{fantasize}{\textsc{fantasize}}
        \stageval{$\mathcal{M}_t$, $\bm X_t$, $\mathcal{D}_{t-1}$}:\\
        \Indp
        $y^*_{t-1} = \max_{(x, y)\in \mathcal{D}_{t-1}} y$\\
        \Return $\mathbb{E}_{y\sim \mathcal{M}_t(\bm X_t)}\bigl[(y - y^*_{t-1})^+ \bigr]$
        \BlankLine
        \Indm
        \step{$\alpha_t$, $\mathcal{M}_t$, $\bm X_{t:k}$, $\bm Z_{t:k}$, $\mathcal{D}_{t-1}$}:\\
        \Indp
        $\alpha_{t+1}$ = $\alpha_t$ + \stageval{$\mathcal{M}_t$, $\bm X_t$, $\mathcal{D}_{t-1}$}\\
        \If{$t = k-1$}{
            \Return $\alpha_{t+1}$
        }
            $Y_t$ = \correlate{$\mathcal{M}_t(\bm X_t)$, $\bm Z_t$}\\
            $\mathcal{M}_{t+1}$ = \fantasize{$\mathcal{M}_t$, $\bm X_t$, $\bm Y_t$}\\
            $\mathcal{D}_{t} = \mathcal{D}_{t-1} \cup \{(\bm X_t, \bm Y_t)\}$\\
            \Return \step{$\alpha_{t+1}$, $\mathcal{M}_{t+1}$, $\bm X_{t+1:k}$, $\bm Z_{t+1:k}$, $\mathcal{D}_t$}\\
        
        \caption{Multi-Step Tree Evaluation}
        \label{algo:OST:FullAlgo}
        \end{algorithm}
    \caption{
    The recursive procedure for evaluating multi-step trees by repeatedly sampling from the posterior
     (\texttt{\textsc{correlate}}), conditioning (\texttt{\textsc{fantasize}}), 
    and evaluating stage-values (\texttt{\textsc{value}}). }
    \end{minipage}
\end{figure*}
\subsection{Multi-Step Fantasies and Scalability}
Our other core insight is that the different levels of the lookahead tree can be represented by the batch dimensions of batched \gp models; this allows us to exclusively use batched linear algebra tensor operations that heavily exploit parallelization and hardware acceleration for our computations. 
This optimized implementation is crucial in order to scale to non-trivial multi-step problems.
Algorithm~\ref{algo:OST:FullAlgo} 
shows our recursive implementation of~\eqref{eq:one-shot-objective}.\footnote{Here $\mathcal{M}_t(\bm x_t)$ denotes the posterior of the model $\mathcal{M}_t$ evaluate at $\bm x_t$, and $\bm X_{t:k} := \{\bm x_i\}_{i=t}^k$ and $\bm Z_{t:k} := \{\bm z_i\}_{i=t}^k$ are collections of decision variables and base samples for lookahead steps~$t$ through~$k$, respectively. \mbox{\textsc{correlate}($\mathcal{M}_t(\bm X_t)$, $\bm Z_t$)} generates fantasy samples by correlating the base samples $\bm Z_t$ via the model posterior $\mathcal{M}_t(\bm X_t)$, and \mbox{\textsc{fantasize}($\mathcal{M}_t$, $\bm X_t$, $\bm Y_t$)} produces a new fantasy model (with an additional batch dimension) by conditioning on the fantasized observations.
To compute $k$-step one-shot lookahead conditional on base samples $\bm Z_{0:k}$ at decision variables $\bm X_{0:k}$, we simply need to call \textsc{Step}($0$, $\mathcal{M}$, $\bm X_{0:k}$, $\bm Z_{0:k}$, $\mathcal{D}$).}
Using reparameterization, we retain the dependence of the value functions in all stages on $x, \mathbf{x}_2, \dots, \mathbf{x}_k$, and can auto-differentiate through Algorithm~\ref{algo:OST:FullAlgo}. 


%

%

Figure~\ref{fig:ParallelHardware:FastFantasies:WallTimes2} compares the overall wall time (on a logarithmic scale) for constructing fantasy models and performing posterior inference, for both standard and fast fantasy implementations. Fast fantasies achieve substantial speedups with growing model size and number of fantasies (up to 16x for $n=1024$, $m=128$). In Appendix~\ref{appdx:sec:FastFantasies}, we show that speedups on a \acro{GPU} are even higher. 

\section{Special Instances of Multi-Step Trees}
\label{sec:pseudo}
The general one-shot optimization problem is of dimension $d + d \sum_{t=1}^{k}\prod_{i=1}^t m_i$,
which grows exponentially in the lookahead horizon~$k$. Therefore, in practice we are limited to relatively small horizons $k$ (nevertheless, we are able to show experimental results for up to $k=4$, which has never been considered in the literature). In this section, we describe two alternative approaches that have dimension only linear in~$k$ and have the potential to be even more scalable.

\textbf{Multi-Step Path.} Suppose only a single path is allowed in each subtree rooted at each fantasy sample $y^{j_1}, j_1=1,\dots, m_1$, that is, let $m_t=1$ for $t\geq 2$, then the number of variables is linear in $k$ and $m_1$. An even more extreme case is further setting $m_1=1$, that is, we assume there is only one possible future path. 
When Gauss-Hermite (\gh) quadrature rules are used (one sample is exactly the mean of the Gaussian), this approach has a strong connection with \emph{certainty equivalent control} \citep{bertsekas2017dynamic}, an approximation technique for solving optimal control problems.
It also relates to some of the notable batch construction heuristics such as Kriging Believer \citep{ginsbourger2010kriging}, or \acro{GP-BUCB} \citep{desautels2014parallelizing}, where one fantasizes (or ``hallucinates'' as is called in \cite{desautels2014parallelizing}) the posterior mean and continue simulating future steps as if it were the actual observed value. 
We will see that this degenerate tree can work surprisingly well in practice.


\textbf{Non-Adaptive Approximation.}
If we approximate the adaptive decisions after the current step by non-adaptive decisions, 
the \emph{one-shot} optimization would be
\begin{gather}
   \SwapAboveDisplaySkip
    \max_{x, X^{(1)}, \dots, X^{(m_1)} } 
    v_1(x\given \data) 
    + 
    \frac{1}{m_1} \sum_{i=1}^{m_1} V^{k-1}_1(X^{(i)} \given \data^{(i)}_1),  \label{eq:one-shot-ensbo}
\end{gather}
where we replaced the adaptive value function $v_{k-1}$ by the one-step batch value function $V^{k-1}_1$ with batch size $k-1$, i.e., $|X| = k-1$.  
The dimension of \eqref{eq:one-shot-ensbo} is $d + m_1 (k-1) d$.   
Since non-adaptive expected utility is no greater than the adaptive expected utility, \eqref{eq:one-shot-ensbo} is a lower bound of the adaptive expected utility. 
Such non-adaptive approximation is actually a proven idea for \emph{efficient nonmyopic search} (\ens) \citep{jiang2017efficient, jiang2018efficient}, a problem setting closely related to \bo.  
We refer to~\eqref{eq:one-shot-ensbo} as \eno, for \emph{efficient nonmyopic optimization}. 
See Appendix \ref{app:lbub} for further discussions of these two special instances.

\begin{table}[th]
    \centering
    \caption{Summary and comparison of the nonmyopic approaches discussed in this paper. 
    Notation: for \glasses, $X_g$ is a heuristically constructed batch; note it does not depend on $y$; for rollout, $r_1(x\given \data) = v_1(x \given \data)$, 
    and we assume $\pi(\data_1) = \argmax_x v_1(x \given \data_1)$ is the base policy for a meaningful comparison. Recall $\data_1 = \data \cup \{(x, y)\}$.
    The relationships are due to (1) non-adaptive expected utility is a lower bound on adaptive expected utility, and (2) $\E[\max \cdot] \ge \max \E[\cdot]$.
    }
    \label{table:comparison_of_bounds}
    \def\arraystretch{1.3}
    \vspace{1ex}
\begin{tabular}{ll}
\toprule
Method  &   Acquisition Function    \\ 
\hline
multi-step (ours)  & $v_1(x\given \data) + \E_y [  \textstyle{  \max_{x'} } v_{k-1}(x' \given \data_1) ]$    \\
    \eno (ours) & $v_1(x\given \data) + \E_y [ \textstyle{  \max_X } V^{k-1}_1(X \given \data_1) ]$  \\
    \binoc \cite{jiang2019binoculars} & $v_1(x\given \data) + \textstyle{ \max_X } \E_y [ V^{k-1}_1(X \given \data_1) ]$   \\
    \glasses \cite{gonzalez2016glasses} & $v_1(x\given \data) + \E_y [ V^{k-1}_1(X_g \given \data_1) ]$    \\
    rollout \cite{lam2016bayesian} & $r_k(x\given \data) = r_1(x\given\data) + \E_y[ r_{k-1}(\pi(\data_1) \given \data_1) ]$  \\ 
    two-step \cite{wu2019practical} & $v_1(x\given \data) + \E_y [  \textstyle{  \max_{x'} } v_{1}(x' \given \data_1) ]$   \\
    one-step \cite{movckus1975bayesian}   & $v_1(x\given \data) + 0 $   \\ 
        \hline 
    \multirow{2}{*}{relationships (when $k\ge2$)}  
                & {multi-step $\ge$ \eno $\ge$ \binoc $\ge$ \glasses $\ge$ one-step;}   \\
                & multi-step $\ge$ rollout $\ge$ two-step $\ge$ one-step;  
                 ~~\eno $\ge$ two-step. \\
            \bottomrule
    \end{tabular}
\end{table}

\section{Related Work}
\label{section:related}
While general multi-step lookahead \bo has never been attempted at the scale we consider here, 
there are several earlier attempts on two-step lookahead \bo \citep{osborne2009gaussian, ginsbourger2010towards}. The most closely related work is a recent development that made gradient-based optimization of two-step \ei possible \citep{wu2019practical}. 
In their approach, maximizers of the second-stage value functions conditioned on each $y$ sample of the first stage are first identified, and then substituted back. 
If certain conditions are satisfied, this function is differentiable and admits unbiased stochastic gradient estimation (envelope theorem).  This method relies on the assumption that the maximizers of the second-stage value functions are \emph{global optima}. This assumption can be unrealistic, especially for high-dimensional problems.
Any violation of this assumption would result in discontinuity of the objective, and differentiation would be problematic.

Rollout is a classical approach in approximate dynamic programming \citep{bertsekas1996neuro,sutton2018reinforcement} and adapted to \bo by \cite{lam2016bayesian,yue2019why}. 
However, rollout estimation of the expected utility is only a lower bound of the true multi-step expected utility, because a \emph{base policy} is evaluated instead of the \emph{optimal policy}.  
Another notable nonmyopic \bo policy is \glasses \citep{gonzalez2016glasses}, which also uses a batch policy to approximate future steps. 
Unlike \eno, they use a heuristic batch instead of the optimal one, and perhaps more crucially, their batch is not adaptive to the sample values of the first stage.
All three methods discussed above share similar repeated, nested optimization procedures for each evaluation of the acquisition function and hence are very expensive to optimize.

Recently \cite{jiang2019binoculars} proposed an 
efficient nonmyopic approach called \binoc, where a point from the optimal batch is selected at random.
This heuristic is justified by the fact that any point in the batch maximizes a lower bound of the true expected utility that is tighter than \glasses. 
We summarize all the methods discussed in this paper in Table~\ref{table:comparison_of_bounds}, in which we also present comparisons of the tightness of the lower bounds.
Note that ``multi-step path'' can be considered a noisy version of ``multi-step''. 

\begin{table}[t]
    \centering
    \caption{Average \gap and time (seconds) per iteration. We run 100  repeats for each method and each function with $2d$ random initial observations of the function. 
    Bold: best, blue: not significantly worse than the best under paired one-sided sign rank test with $\alpha=0.05$.}
    \label{table:synthetic_results}
    \vspace{1ex}
    \small
\begin{tabular}{lllllllll}
\toprule
&{EI} & {ETS} & {12.EI.s} & {2-step} & {3-step} & {4-step} & {4-path} & {12-\eno}\\\hline
{eggholder} & 0.627   &  0.647   &  \textbf{0.736  } &  0.478   &  0.536   &  0.577   &  0.567   &  0.661   \\  
{dropwave} & 0.429   &  0.585   &  0.606   &  0.545   &  0.600   &  0.635   &  \textbf{0.731  } &  0.673   \\  
{shubert} & 0.376   &  \textit{\textcolor{blue}{0.487  }} &  \textit{\textcolor{blue}{0.515  }} &  0.476   &  \textit{\textcolor{blue}{0.507  }} &  \textbf{0.562  } &  \textit{\textcolor{blue}{0.560  }} &  \textit{\textcolor{blue}{0.494  }} \\  
{rastrigin4} & 0.816   &  0.495   &  0.790   &  \textbf{0.851  } &  \textit{\textcolor{blue}{0.821  }} &  \textit{\textcolor{blue}{0.826  }} &  \textit{\textcolor{blue}{0.837  }} &  \textit{\textcolor{blue}{0.837  }} \\  
{ackley2} & 0.808   &  0.856   &  \textit{\textcolor{blue}{0.902  }} &  \textit{\textcolor{blue}{0.870  }} &  \textit{\textcolor{blue}{0.895  }} &  \textit{\textcolor{blue}{0.888  }} &  \textbf{0.931  } &  0.847   \\  
{ackley5} & 0.576   &  0.516   &  0.703   &  0.786   &  0.793   &  0.804   &  \textbf{0.875  } &  \textit{\textcolor{blue}{0.856  }} \\  
{bukin} & 0.841   &  0.843   &  0.842   &  \textbf{0.862  } &  \textit{\textcolor{blue}{0.862  }} &  \textit{\textcolor{blue}{0.861  }} &  \textit{\textcolor{blue}{0.852  }} &  0.836   \\  
{shekel5} & 0.349   &  0.132   &  0.496   &  \textit{\textcolor{blue}{0.827  }} &  \textbf{0.856  } &  \textit{\textcolor{blue}{0.847  }} &  0.718   &  0.799   \\  
{shekel7} & 0.363   &  0.159   &  0.506   &  \textit{\textcolor{blue}{0.825  }} &  \textit{\textcolor{blue}{0.850  }} &  0.775   &  0.776   &  \textbf{0.866  } \\  
\bottomrule
{Average} & 0.576   &  0.524   &  0.677   &  0.725   &  \textit{\textcolor{blue}{0.747  }} &  \textit{\textcolor{blue}{0.753  }} &  \textit{\textcolor{blue}{0.761  }} &  \textbf{0.763  } \\  
Ave. time & 1.157 & 1949. & 25.74  & 7.163 & 39.53 & 197.7 & 17.50 & 15.61 \\
\bottomrule
\end{tabular}
\end{table}

\section{Experiments}
\label{sec:exper}

We follow the experimental setting of \citep{jiang2019binoculars}, 
and test our algorithms using the same set of synthetic and real benchmark functions. Due to the space limit, we only present the results on the synthetic benchmarks here; more results are given in Appendix \ref{appd:realfunctions}. 
All algorithms are implemented in \botorch \citep{balandat2019botorch}, and we use a \gp with a constant mean and a Mat\'ern $\nicefrac{5}{2}$ \acro{ARD} kernel for \bo. \gp hyperparameters are re-estimated by maximizing the evidence after each iteration. 
For each experiment, we start with $2d$ random observations, and perform $20d$ iterations of \bo; 100 experiments are repeated for each function and each method. 
We measure performance with \gap $=(y_i - y_0)/(y^* - y_0)$. 
All experiments are run on \acro{CPU} Linux machines; each experiment only uses one core.  

\cite{jiang2019binoculars} used nine ``hard'' synthetic functions, motivated by the argument that the advantage of nonmyopic policies are more evident when the function is hard to optimize.
We follow their work and use the same nine functions. 
\cite{jiang2019binoculars} thoroughly compared \binoc with some well known nonmyopic baselines such as rollout and \glasses and demonstrate superior results, so we will focus on comparing with \ei, \binoc and the ``envelope two-step'' (\ets) method \citep{wu2019practical}. We choose the best reported variant 12.\ei.s for \binoc on these functions \citep{jiang2019binoculars},  i.e., first compute an optimal batch of 12 points (maximize \qei), then sample a point from it weighted by the individual \ei values. 
All details can be found in our attached code. 

\begin{figure*}
	\centering
	\begin{subfigure}[b]{.45\textwidth}
		\centering
		\includegraphics[width=\linewidth]{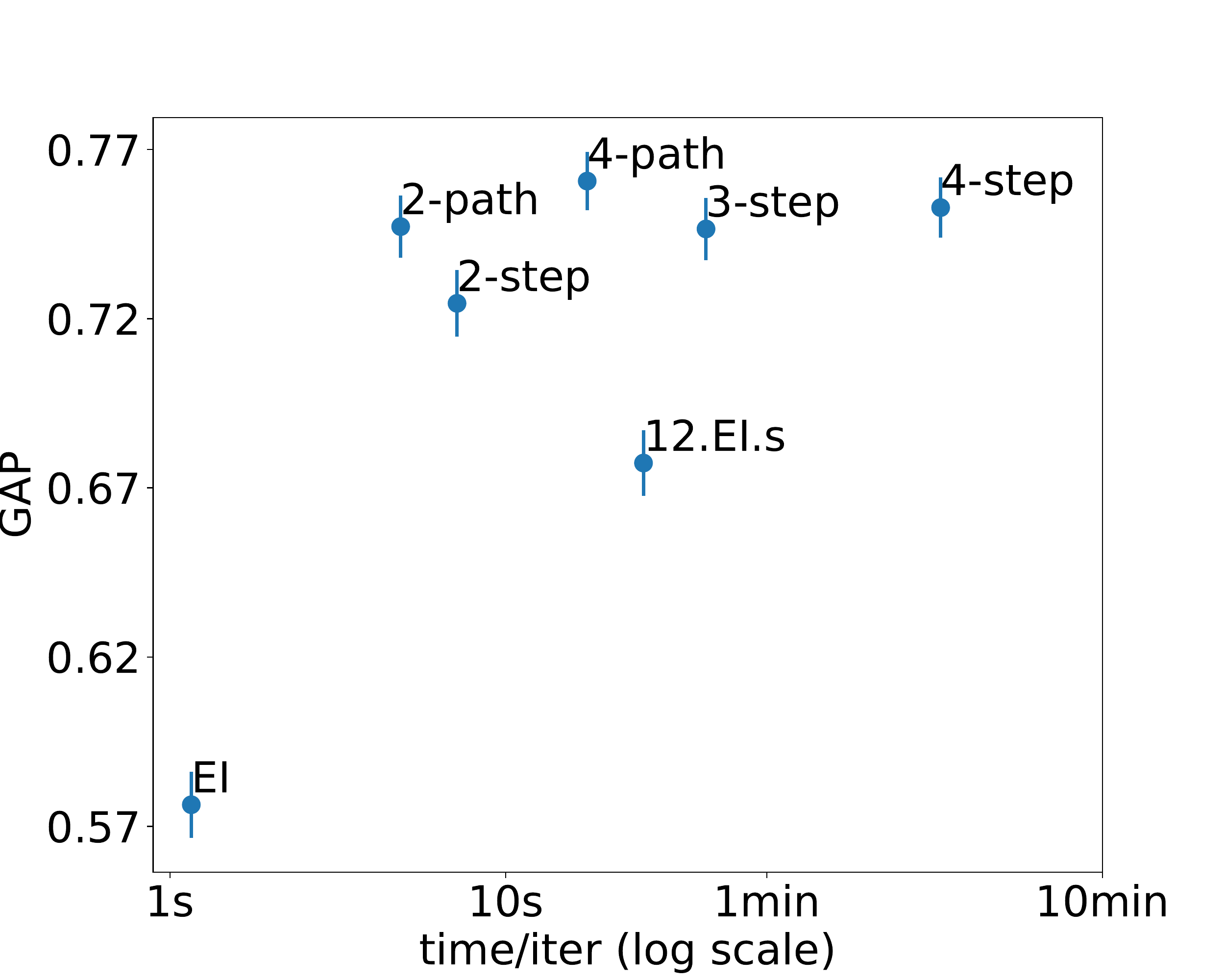}
		\subcaption{}
		\label{fig:synthetic_gap_time}
	\end{subfigure}
	\hspace{0.5cm}
	\begin{subfigure}[b]{.45\textwidth}
		\centering
		\includegraphics[width=\linewidth]{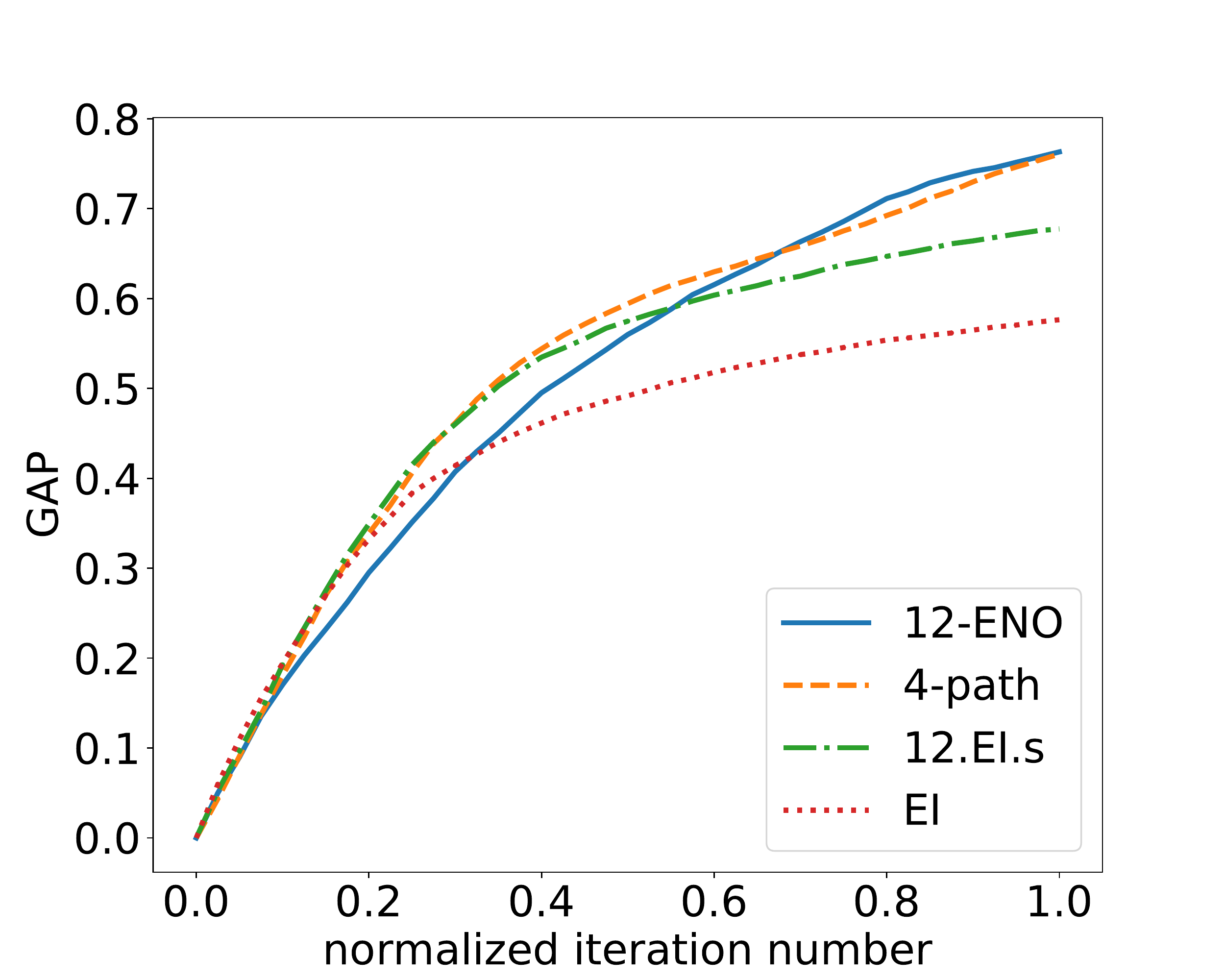}
		\subcaption{}
		\label{fig:nonmyopic_behavior}
	\end{subfigure}
	\caption{(a) Aggregated \gap with error bars vs. time per iteration, averaged over the nine synthetic functions by 100 repeats.
		(b) Aggregated optimization trace: \gap versus number of iterations, normalized into 0-1.}
\end{figure*}

We use the following nomenclature: 
``$k$-step'' means $k$-step lookahead ($k=2,3,4$) with number of \gh samples $m_1=10, m_2=5, m_3=3$. These numbers are heuristically chosen to favor more
accuracy in earlier stages.
``$k$-path'' is the multi-step path variant that uses only one sample for each stage. 
``$k$-\eno'' is \eno using non-adaptive approximation of the future $k-1$ steps with a $q$-\ei batch with $q=k-1$.
The average \gap at terminal and average time (in seconds) per iteration for all methods are presented in Table \ref{table:synthetic_results}.
The average over all functions are also plotted in Figure \ref{fig:synthetic_gap_time}. 
Some entries are omitted from the table and plots for better presentation. We summarize some main takeaways below.

\textbf{Baselines.} First, we see from Table \ref{table:synthetic_results} \ets outperforms \ei on $\nicefrac{5}{9}$ of the functions, but on average is worse than \ei, especially on the two shekel functions, and the average time/iter is over 30min. 
Our results for 12.\ei.s and \ei closely match those reported in \citep{jiang2019binoculars}. 

\textbf{One-Shot Multi-Step.} We see from Figure \ref{fig:synthetic_gap_time} 
\emph{2,3,4-step outperform all baselines by a large margin}.
We also see diminishing returns with increasing horizon. 
In particular, performance appears to stop improving significantly beyond 3-step.    
It is not clear at this point whether this is because we are not optimizing the increasingly complex multi-step objective well enough or 
if additional lookahead means increasing reliance on the model being accurate, which is often not the case in practice \citep{yue2019why}. 

\textbf{One-Shot Pseudo Multi-Step.}
Note that the average time/iteration grows exponentially when looking more steps ahead if we use multiple fantasy samples for each stage.
In Figure \ref{fig:synthetic_gap_time}, we see 2- and 4-path produce similar results in significantly less time, {suggesting that a noisy version of the multi-step tree is a reasonable alternative.}
The average \gap and time for 2,3,4-path are: \gap= 0.747, 0.750, 0.761, time = 4.87s, 10.4s, 17.5s. We also run 12-\eno. $k=12$ is chosen to match with 12.\ei.s. The \gap and time/iteration are very close to 4-path.
More interestingly, it exhibits more evident \emph{nonmyopic behavior} than other nonmyopic methods shown in Figure \ref{fig:nonmyopic_behavior}: 
12-\eno first underperforms, but catches up at some point, and finally outperforms, the less myopic methods. 
Similar behavior has been consistently observed in efficient nonmyopic active search \citep{jiang2017efficient, jiang2018efficient}.
All plots for the individual functions are given in Appendix \ref{appd:syntheticfunctions}. {Note that these pseudo multi-step methods have objectives that are much lower dimensional and hence easier to optimize --- this may partly contribute to their effectiveness.}

\section{Conclusion}
\label{section:conclusion}
General multi-step lookahead Bayesian optimization is a notoriously hard problem. 
We provided the first efficient implementation based on a simple idea: jointly optimize all the decision variables in a multi-step tree in \emph{one-shot}, instead of naively computing the nested expectation and maximization. 
Our implementation relies on fast, differentiable fantasization, highly batched and vectorized recursive sampling and conditioning of Gaussian processes, and auto-differentiation. 
Results on a wide range of benchmarks demonstrate its high efficiency and optimization performance. 
We also find two special cases, fantasizing posterior mean and non-adaptive approximation of future decisions, work as well with much less time. 
An interesting future endeavor is to investigate the application of our framework to other problems, such as Bayesian quadrature \citep{jiang2019binoculars}.
\clearpage
\section*{Broader Impact}
The central concern of this investigation is Bayesian optimization of an expensive-to-evaluate objective function. As is standard in this body of literature, our proposed algorithms make minimal assumptions about the objective, effectively treating it as a ``black box.'' This abstraction is mathematically convenient but ignores ethical issues related to the chosen objective.  Traditionally, Bayesian optimization has been used for a variety of applications, including materials design and drug discovery \citep{frazier2018tutorial}, and could have future applications to algorithmic fairness. We anticipate that our methods will be utilized in these reasonable applications, but there is nothing inherent to this work, and Bayesian optimization as a field more broadly, that preclude the possibility of optimizing a nefarious or at least ethically complicated objective.

\bibliography{main}
\bibliographystyle{abbrvnat}

\newpage
\appendix

\begin{center}
\hrule height 4pt
\vskip 0.25in
\vskip -\parskip
    {\LARGE\bf  Appendix to:\\[2ex] \papertitle}
\vskip 0.29in
\vskip -\parskip
\hrule height 1pt
\vskip 0.2in%
\end{center}

\section{Proof of Proposition 1}
If we instantiate a different version of $x_t, t=2,\dots, k$ for each realization of $\data_{t-1}^{j_1\dots j_{t-1}}$, 
we can move the maximizations outside of the sums: 
\begin{align}
\begin{split}
\label{eq:OST:MST:MCapprox}
    \bar{v}_k(x\given \data) 
    = 
    v_1(x \given \data) 
    +
   \max_{\mathbf{x}_2, \mathbf{x}_3, \dots}
   \biggl\{ &
   \frac{1}{m_1}\sum_{j_1=1}^{m_1} v_1(x_2^{j_1} \given \data_1^{j_1}) + \\[-2ex]
    &\qquad\qquad\frac{1}{m_1 m_2}\sum_{j_1=1}^{m_1}\sum_{j_2=1}^{m_2}  v_1(x_3^{j_1 j_2} \given \data_2^{j_1 j_2}) +
    \cdots  \biggr\}.
\end{split}
\end{align}

We make the following general observation. Let $G(x) := \max_{\tilde{x} \in \tilde{\Omega}} g(x,\tilde{x})$ for some $g: \Omega \times \tilde{\Omega} \rightarrow \mathbb{R}$. If $(x^*, \tilde{x}^*) \in \argmax_{(x, \tilde{x}) \in \Omega \times \tilde{\Omega}} g(x,\tilde{x})$, then $x^* \in \argmax_{x \in \Omega} G(x)$. The result follows directly by viewing $v_k(x\given\data)$ as $G$ and the objective on the right-hand-side of (\ref{eq:one-shot-objective}) as $g$.

\section{One-Shot Optimization of Lower and Upper Bounds}
\label{app:lbub}
As described in the main text, the non-adaptive approximation for pseudo multi-step lookahead corresponds to a one-shot optimization of a lower bound on Bellman's equation.  Here we show this from a different perspective and also demonstrate that the multi-step path approach can be viewed as one-shot optimization of an upper bound depending on the implementation.

We can generate a lower-bound on the reparameterized \eqref{eq:recursive_max_and_exp} by moving maxes outside expectations:
\begin{align}
    v_k(x\given \data) 
    &\geq  
    v_1(x \given \data) 
    +
    \E_z \Big[\max_{x_2, x_3,\ldots, x_k} \E_{z_2, z_3 \ldots z_{k-1}} \sum_{i=2}^{k} v_1(x_i \given \data_{i-1})\Big] \\
    &=  v_1(x \given \data) + \E_z[ \max_{X} V^{k-1}_1(X \given \data_1 ) ],
    \label{eq:lower_bound}
\end{align}
where we recognize the second line as the objective for the non-adaptive approximation.  

We similarly generate an upper-bound on the reparameterized \eqref{eq:recursive_max_and_exp} by moving maximizations inside the expectations as follows:
\begin{align}
    v_k(x\given \data) 
    \leq  
    v_1(x \given \data) 
    +
    \E_{z,z_2, z_3,\ldots,z_{k-1}} \Big[\max_{x_2, x_3,\ldots, x_k} \sum_{i=2}^{k} v_1(x_i \given \data_{i-1})\Big].
    \label{eq:upper_bound}
\end{align}
This equation includes an expectation of paths of base samples $z, z_2, z_3,\ldots,z_{k-1}$ rooted at $x$.  Consider one-shot optimization of the right-hand side of this equation with $m$ base sample paths:
\begin{align}
    x^*, \mathbf{x}^*_2, \mathbf{x}^*_3, \dots, \mathbf{x}^*_k
    =
    \argmax_{x, \mathbf{x}_2, \mathbf{x}_3, \dots, \mathbf{x}_k} \Biggl\{
    v_1(x \given \data) 
    +
   \frac{1}{m}\sum_{j_1=1}^{m} \sum_{i=2}^k v_1(x_i^{j_1} \given \data_{i-1}^{j_1})\Biggr\}. \label{eq:one-shot-objective-bound}
\end{align}
Comparing to \eqref{eq:one-shot-objective}, the two expressions are identical when $m_1=m$, and $m_2=m_3=m_4\ldots=m_{k-1}=1$. In the main text, our approach through Gauss-Hermite quadrature produces base samples $z_2=z_3=z_4=\ldots z_{k-1}=0$, which does not correspond to an approximation of \eqref{eq:upper_bound}.  However, if we actually used Monte-Carlo (or quasi-Monte Carlo) base samples for $z$, the one-shot optimization for multi-step paths would correspond to \eqref{eq:upper_bound}.  

One can also derive analogous sample-specific bounds on one-shot trees defined by a given set of base samples.  Forcing decision variables at the same level in the tree to be identical is the fixed sample equivalent of moving a max outside an expectation, and enforces a lower-bound on the specific one-shot tree.  Moving a max inside the expectation also has a fixed sample equivalent: splitting the decision variable based on each base sample for that expectation, replicating the descendant tree including the base samples and decision variables, and then allowing all of these additional decision variables to be independently optimized produces an upper-bound on the tree.

\section{Implementation Details}
\label{subsec:OST:Implementation}

\textbf{Optimization.} 
Because of its relatively high dimensionality, the multi-step lookahead acquisition function can be challenging to optimize. Our differentiable one-shot optimization strategy enables us to employ deterministic (quasi-) higher-order optimization algorithms, which achieve faster convergence rates compared to the commonly used stochastic first-order methods~\citep{balandat2019botorch}.
This is in contrast to zeroth-order optimization of most existing nonmyopic acquisition functions such as rollout and \glasses.
To avoid computing Hessians via auto-differentiation, we use \acro{L-BFGS-B}, a quasi-second order method, in combination with a random restarting strategy to optimize~\eqref{eq:one-shot-objective}.

\textbf{Warm-start Initialization.}
In our empirical investigation, we have found that careful initialization of the multi-step optimization is crucial. To this end, we developed an advanced warm-starting strategy inspired by homotopy methods that re-uses the solution from the previous iteration (see Appendix~\ref{appdx:sec:WarmStart}). 
Using this strategy dramatically improves the \bo performance relative to a naive optimization strategy that does not use previous solutions. 


\textbf{Gauss-Hermite Quadrature.}
Instead of performing \acro{MC} integration, we can also use Gauss-Hermite (\gh) quadrature rule to draw samples for approximating the expectations in each stage~\citep{lam2016bayesian, wu2019practical}. In this case, when using a single sample as in the case of ``multi-step path'', the sample value is always the mean of the Gaussian distribution. 

We implemented multi-step fast fantasies in \gpytorch \citep{gardner2018gpytorch}, and the multi-step lookahead acquisition function in \botorch \citep{balandat2019botorch}. 
Our code is included as part of this submission, and will be made public under an open-source license.

\section{Warm-Start Initialization Strategy for Multi-Step Trees}
\label{appdx:sec:WarmStart}

Warm-starting is an established method to accelerate optimization algorithms based on the solution or partial solution of a similar or related problem, specifically in case the problem structure remains fixed and only the parameters of the problem change. This is exactly the situation we find ourselves in when optimizing acquisition functions for Bayesian optimization more generally, and optimizing multi-step lookahead trees more specifically. 

Since the multi-step tree represents a scenario tree, one intuitive way of warm-starting the optimization is to identify that branch originating at the root of the tree whose fantasy sample is closest to the value actually observed when evaluating the suggested candidate on the true function. This sub-tree is that hypothesized solution that is most closely in line with what actually happened. One can then use this sub-tree of the previous solution as a way of initializing the optimization.

Let $\bm X_{0:k}^* := \{\bm x_i^*\}_{i=0}^k$ be the solution tree of the random restart problem that resulted in the maximal acquisition value in the previous iteration. For our restart strategy, we add random perturbations to the different fantasy solutions, increasing the variance as we move down the layers in the optimization tree (depth component that captures increasing uncertainty the longer we look ahead), and increasing variance overall (breadth component that encourages diversity of the initial conditions to achieve coverage of the domain). Concretely, supposing w.l.o.g.\ that $\mathcal{X} = [0, 1]$, we generate $N$ initial conditions $\bm X_{0:k}^1, \dotsc, \bm X_{0:k}^{N}$ as 
\begin{align}
    \bm x^r_i = (1-\gamma_r) \Bigl( (1 - \eta_i) \bm x^*_i + \eta_i \beta^r_i \Bigr) + \gamma_r u^r_i,
\end{align}
with $\beta^r_i$ and $u^r_i$ of the same shape as $\bm x^*_i$, with individual elements drawn i.i.d.\ as $\beta^r_i \sim \text{Beta}(1, 3)$ and $u^r_i \sim U[0, 1]$ for all $r, i$, and $\gamma_1 < \dotsc < \gamma_{N}$ and $\eta_0 < \dotsc < \eta_k$ are hyperparameters (in practice, a linear spacing works well).

\section{Fast Multi-Step Fantasies}
\label{appdx:sec:FastFantasies}

Our ability to solve true multi-step lookahead problems efficiently is made feasible by linear algebra insights and careful use of efficient batched computation on modern parallelizable hardware.

\subsection{Fast Cache Updates}
\label{appdx:subsec:FastFantasies:FastCache}

If $R$ were a full Cholesky decomposition of $\tilde{K}_{\X\X}$, it could be updated in $\bigo{n^2}$ time. This is advantageous, because computing the Cholesky decomposition required $\bigo{n^3}$ time. However, for dense matrices, the \LOVE cache requires only $\bigo{n^2r}$ time to compute. Therefore, to use it usefully for multi-step lookahead, we must demonstrate that it can be updated in $o(n^2)$ time.

Suppose we add~$q$ rows and columns to $\tilde{K}_{\X\X}$ (e.g. by fantasizing at a set~$\mathbf{x} \in \mathbb{R}^{q \times d}$ of candidate points) to get:
\begin{equation*}
    \left[\begin{array}{cc}
    \tilde{K}_{\X\!\X} & U \\
    U^{\top} & S
    \end{array}\right],
\end{equation*}
where $U \in \mathbb{R}^{n \times q}$ and $S \in \mathbb{R}^{q \times q}$. One approach to updating the decomposition with~$q$ added rows and columns is to correspondingly add~$q$ rows and columns to the update. By enforcing a lower triangular decomposition without loss of generality, this leads to the following block equation:
\begin{equation*}
    \left[\begin{array}{cc}
    \tilde{K}_{\X\X} & U \\
    U^{\top} & S
    \end{array}\right] \approx \left[\begin{array}{cc}
    R & 0 \\
    L_{12} & L_{22} 
    \end{array}\right]\left[\begin{array}{cc}
    R & 0 \\
    L_{12} & L_{22} 
    \end{array}\right]^{\top}
\end{equation*}
From this block equation, one can derive the following system of equations:
\begin{subequations}
\begin{align*}
    \tilde{K}_{\X\X} &= RR^{\top} \\
    U &= RL_{12}^{\top} \\
    S &= L_{12}L_{12}^{\top} + L_{22}L_{22}^{\top}
\end{align*}
\end{subequations}
To compute $L_{12}^{\top}$, one computes $R^{-1}U$. Since in the \LOVE case $R$ is rectangular, this must instead be done by solving a least squares problem. To compute $L_{22}$, one forms $S - L_{12}L_{12}^{\top}$ and decompose it.

\textbf{Time Complexity.} In the special case of updating with a single point, ($q = 1$), $U \in \mathbb{R}^{n}$ and $S \in \mathbb{R}$. 
Therefore, computing $L_{12}$ and $L_{22}$ requires the time of computing a single \LOVE variance ($R^{-1}U$) and then taking the square root of a scalar ($S - L_{12}L_{12}^{\top}$). In the general case, with a cached pseudoinverse for~$R$, the total time complexity of the update is dominated by the multiplication $R^{-1}U$ assuming~$q$ is small relative to~$n$, and this takes $\bigo{nrq}$ time. Note that this is a substantial improvement over the $\bigo{n^2q}$ time that would be required by performing rank 1 Cholesky. If in each of $k$ steps of lookahead we are to condition on $m$ samples at $q$ locations, the total running time required for posterior updates is $\bigo{nrqm^{k-1}}$.

\textbf{Updating the Inverse.} The discussion above illustrates how to update a cache~$R$ to a cache that incorporates~$q$ new data points. In addition, one would like to cheaply update a cache for~$R^{-1}$ without having to \QR decompose the full new cache. From inspecting a linear systems / least squares problems of the form
\begin{equation}
    \left[\begin{array}{cc}
    R & 0 \\
    L_{12} & L_{22} 
    \end{array}\right]\left[\begin{array}{c}
    x \\ y
    \end{array}\right] = \left[\begin{array}{c}
    b \\ c
    \end{array}\right]
\end{equation}
one can find that $x = R^{-1}b$ and $y = L_{22}^{-1}(c - L_{12}R^{-1}b)$.
Therefore, an update to the (pseudo-)inverse is given by:
\begin{equation}
     \left[\begin{array}{cc}
    R & 0 \\
    L_{12} & L_{22} 
    \end{array}\right]^{-1} = \left[\begin{array}{cc}
    R^{-1} & 0 \\
    -L_{22}^{-1}L_{12}R^{-1} & L_{22}^{-1} 
    \end{array}\right].
\end{equation}

\textbf{Practical Considerations.} Suppose one wanted to compute cache updates at~$J$ different sets of points~$\mathbf{x}^{(1)}, \dotsc \mathbf{x}^{(J)}$, each of size~$q$. For efficiency, one can store a single copy of the pre-computed cache $R^{-1}$ and just compute the update term $P_k :=\bigl[-L_{22,(j)}^{-1}L_{12,(j)}R^{-1} \;\;\; L_{22,(j)}^{-1} \bigr]$ for each of the~$\mathbf{x}^{(j)}$. To perform a solve with the full matrix above and an $n + q$ size vector $\bv$, one can now compute $R^{-1}\bv[:\!n]$, $P\bv$, and concatenate the two. Therefore, an efficient implementation of this scheme is to compute a batch matrix $P \in \mathbb{R}^{J \times q \times (r + q)}$ containing $P_1, \dotsc, P_K$ given a single copy of $R^{-1}$. This allows handling multiple input points by converting a single \gpytorch \gp model to a batch-mode \gp with a batched cache, all while still only storing a single copy of~$R^{-1}$.

\subsection{Multi-Step Fast Fantasies}
\label{appdx:subsec:FastFantasies:MultiStep}

\textbf{Batched Models.} 
For our work, we extend the above cache-updating scheme from \gpytorch to the multi-step lookahead case, in which we need to fantasize from previously fantasized models. To this end, we employ \gpytorch models with multiple batch dimensions. These models are a natural way of representing multi-step fantasy models, in which case each batch dimension represents one level in the multi-step tree. Fantasizing from a model then just returns another model with an additional batch dimension, where each batch represents a fantasy model generated using one sample from the current model's posterior. Since this process can also be applied again to the resulting fantasy models, this approach makes it straightforward to implement multi-step fantasy models in a recursive fashion.



\textbf{Efficient Fantasizing.} 
Note that when fantasizing from a model (assuming no batch dimensions for notational simplicity), each fantasy sample~$y_t^{j_i}$ is drawn at the same location~$x_t$. Since the cache does not depend on the sample~$y_t^{j_i}$, we need to compute the cache update only once. We still use a batch mode \gp model to keep track of the fantasized values,\footnote{Making sure not to perform unnecessary copies of the initial data, but instead share memory when possible.} but use a single copy of the updated cache~$\tilde{R}^{-1}$. We utilize PyTorch's tensor broadcasting semantics to automatically perform the appropriate batch operations, while reducing overall memory complexity.

\textbf{Memory Complexity.} 
For multi-step fantasies, using efficient fantasizing means that the cache can be built incrementally in a very memory- and time-efficient fashion.
Suppose we have a (possibly approximate) root decomposition for~$R^{-1}$ of rank $r \leq n$, i.e. $R\in \mathbb{R}^{n \times r}$. The naive approach to fantasizing is to compute an update by adding $q_t$ rows and columns in the~$t$-th step for each fantasy branch. For a $k$-step look-ahead problem, this requires storing a total of  
\begin{align}
\label{eq:sec:ComplexityFF:NaiveFull}
    N_{\text{naive}} = nr + \sum_{t=0}^{k-1} \prod_{\tau=0}^t m_\tau \left(n + \sum_{\tau=0}^t q_\tau\right)\left(r + \sum_{\tau=0}^t q_\tau\right)
\end{align}
entries. For fast fantasies, we re-use the original~$R^{-1}$ component for each update, and in each step broadcast the matrix across all fantasy points, since the posterior variance update is the same. This means storing a total of
\begin{align}
\label{eq:sec:ComplexityFF:FFFull}
    N_{\text{FF}} = nr + \sum_{t=0}^{k-1} \prod_{\tau=0}^{t-1} m_\tau \, q_t\left(r + \sum_{\tau=0}^t q_\tau\right)
\end{align}
entries. If~$q_t = q$ and $m_t = m$ for all~$t$, then this simplifies to  
\begin{subequations}
\label{eq:sec:ComplexityFF:Simplified}
\begin{align}
\label{eq:sec:ComplexityFF:Simplified:Naive}
    N_{\text{naive}} &= nr + \sum_{t=0}^{k-1} m^{t+1} \left(n + (t+1) q\right)\left(r + (t+1) q\right)
    \\
    N_{\text{FF}} &= nr + \sum_{t=0}^{k-1} m^t \, q\left(r + (t+1) q\right)
\end{align}
\end{subequations}
Assuming $r=n$, we have $N_{\text{naive}} = \bigo{m^k(n+kq)^2}$ and $N_{\text{FF}} = \bigo{m^{k-1}q(n+kq)}$.

\begin{figure}[ht]
    \centering
    \includegraphics[width=\textwidth]{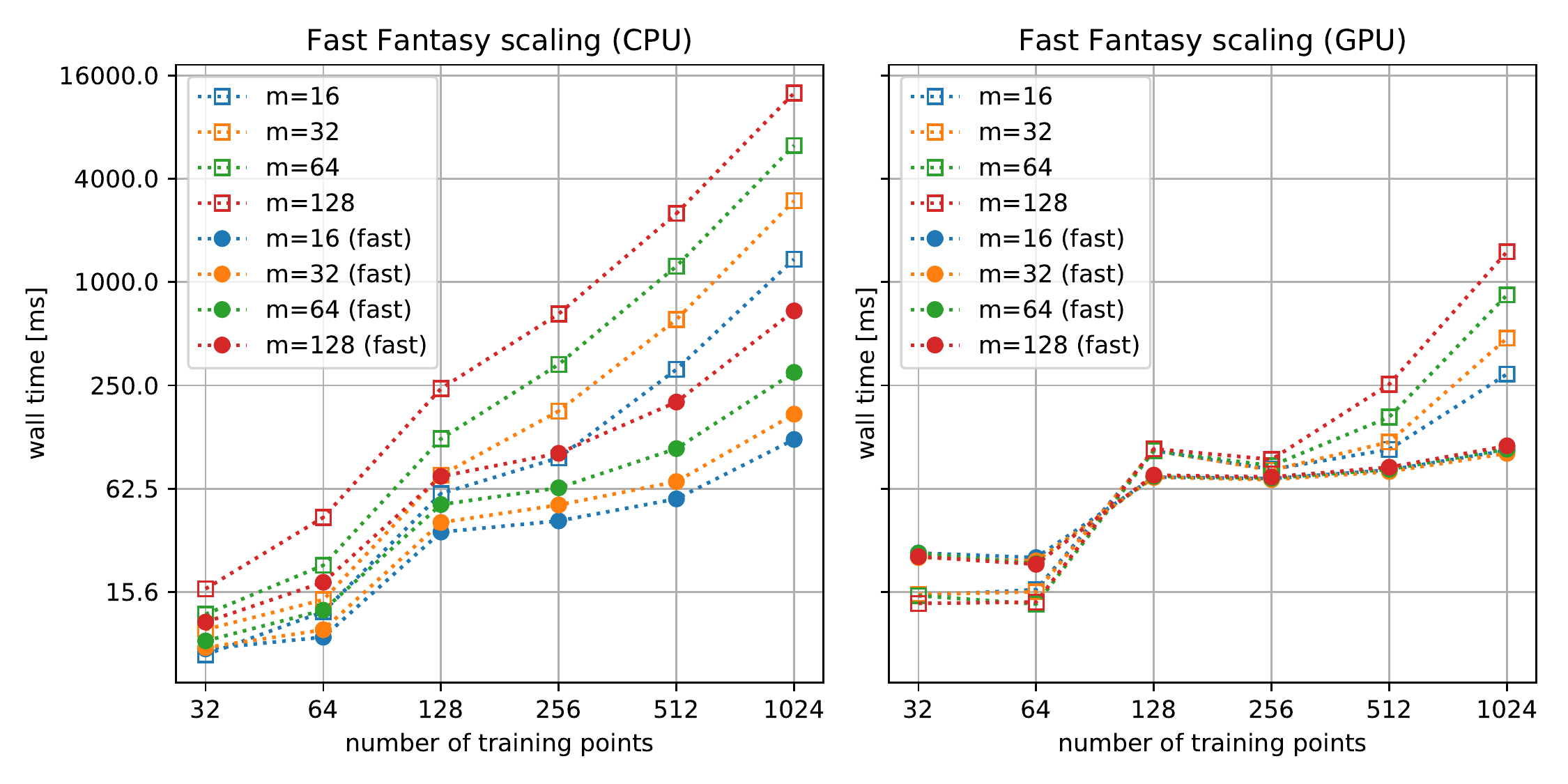}
    \caption{Fast Fantasy wall time comparisons (log-scale). Wall time measures constructing the fantasy model and evaluating its posterior at a single point. Estimated over multiple runs, with variance negligible relative to the mean estimates.
    Results were obtained on an \acro{NVIDIA} Tesla M40 \acro{GPU} - we expect to see even more significant speedups on more modern hardware.}
    \label{fig:ParallelHardware:FastFantasies:WallTimes}
\end{figure}

\textbf{Scalability.}
Figure~\ref{fig:ParallelHardware:FastFantasies:WallTimes} compares the overall wall time (on a logarithmic scale) for constructing fantasy models and performing posterior inference, for both standard and fast fantasy implementations. On  \acro{CPU} fast fantasies are essentially always faster, while on the \acro{GPU} for small models performing full inference is fast enough to outweigh the time required to perform the additional operations needed for performing fast fantasy updates.\footnote{We can also observe interesting behavior on the \acro{GPU}, where inference for 64 training points is faster than for 32 (similarly, for 256 vs. 128). We believe this is due to these sizes working better with the batch dispatch algorithms on the PyTorch backend.}
For larger models we see significant speedups from using fast fantasy models on both \acro{CPU} (up to 22x speedup) and \acro{GPU} (up to 14x speedup).

\section{Results on Real Functions}
\label{appd:realfunctions}
We again use the same set of seven real functions as in \cite{jiang2019binoculars}. 
They are \svm, \lda, logistic regression (LogReg) hyperparameter tuning first introduced in \cite{snoek2012practical},
 neural network tuning on the Boston Housing and Breast Cancer datasets, and active learning of robot pushing first introduced in \cite{wangICML2017b}, and later also used in \cite{malkomes2018automating}. 
 These functions are pre-evaluated on a dense grid.
Log transform of certain dimensions of \svm, \lda, and LogReg are first performed if the original grid is on log scale. 
 We follow \cite{eggensperger2015efficient,eggensperger2018efficient} and use a random forest (\acro{RF}) surrogate model to fit the precomputed grid, and treat the \texttt{predict} function of the trained \acro{RF} model as the target function. 
 We find that the \texttt{RandomForestRegressor} with default parameters in \texttt{scikit-learn}
 can fit the data well, with cross validation $R^2$ mostly over 0.95. 
 A Python notebook is included in our attached code reporting the \acro{RF} fitting results. 
 
\begin{figure*}
	\centering
	\begin{subfigure}[b]{.32\textwidth}
		\centering
		\includegraphics[width=\linewidth]{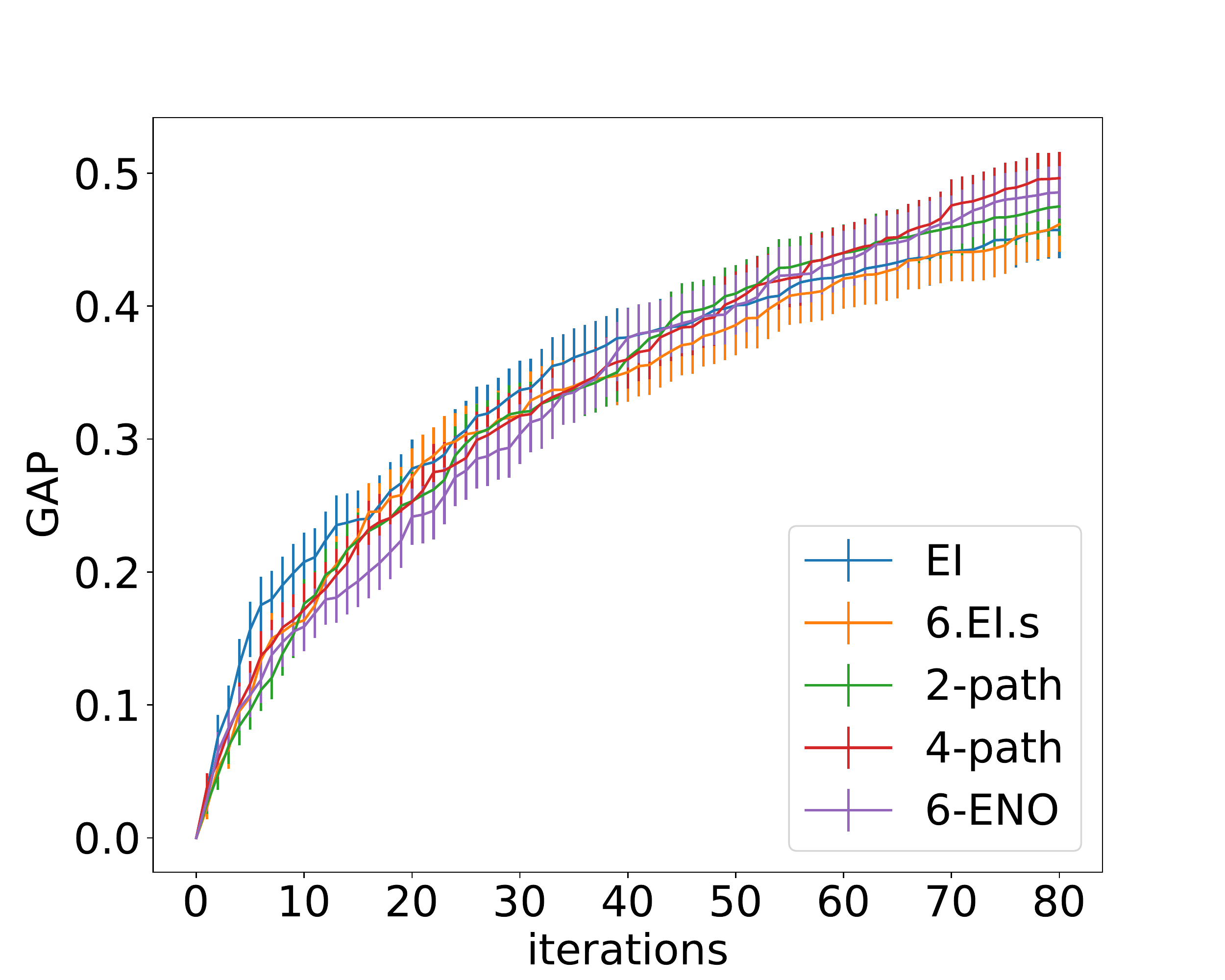}
		\subcaption{NN Boston}
		\label{fig:nn_boston}
	\end{subfigure}
	\begin{subfigure}[b]{.32\textwidth}
		\centering
		\includegraphics[width=\linewidth]{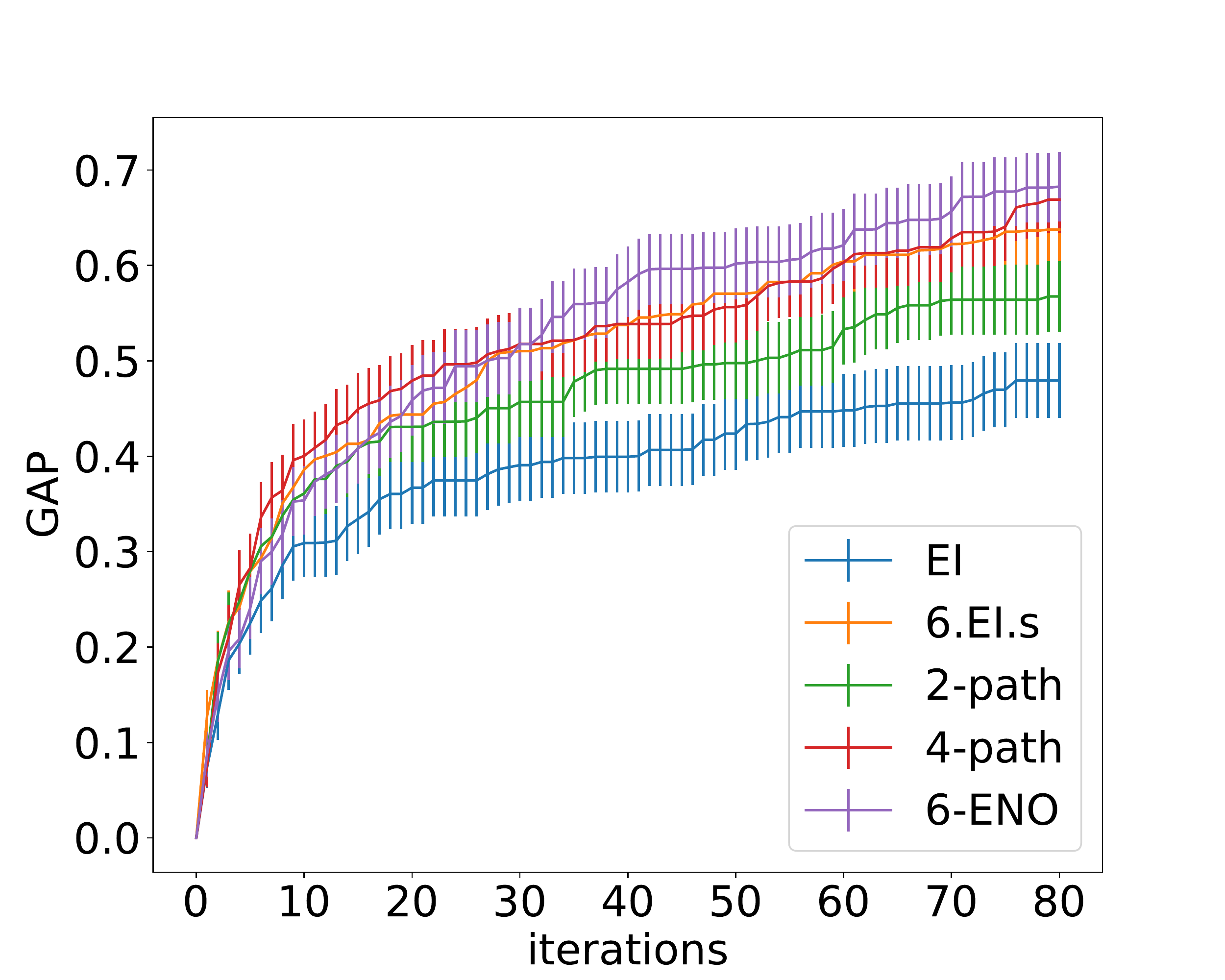}
		\subcaption{NN Cancer}
		\label{fig:nn_cancer}
	\end{subfigure}
	\begin{subfigure}[b]{.32\textwidth}
		\centering
		\includegraphics[width=\linewidth]{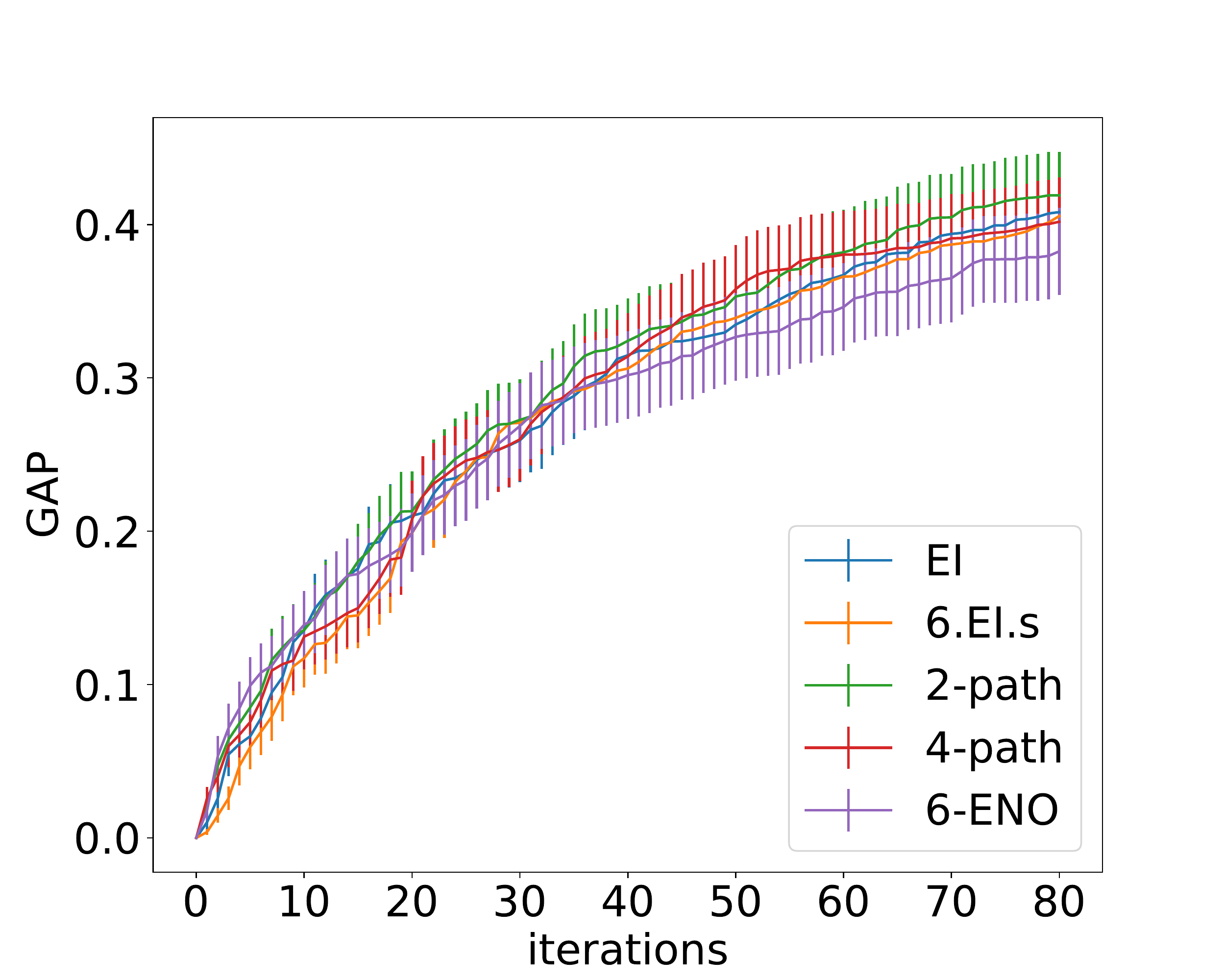}
		\subcaption{Robotpush4}
		\label{fig:robotpush4}
	\end{subfigure}
	\caption{\gap vs. \#iterations on two neural network hyperparameter tuning functions. (a) regression network tuning on the Boston housing dataset.
		(b) classification network tuning on the breast cancer dataset.}
		\label{fig:hard_real_functions}
\end{figure*}

 Table \ref{table:real_function_results} shows the results. The functions are arranged in decreasing order of \ei \gap values. 
 6.\ei.s is the best reported \binoc variant in \cite{jiang2019binoculars} for these functions.
 We only show results for $k$-path ($k=2,3,4$) and 6-\eno. 
 We can see when the function is ``easy'' (e.g., \ei \gap > 0.8), there is almost no difference among all these methods. 
 If we only average over the ``hard'' ones, we see a more consistent and significant pattern as shown in the last row of Table \ref{table:real_function_results}.
 We also plot the \gap curve vs. iterations for the three harder functions in Figure \ref{fig:hard_real_functions}.
 Note the improvement of our method over baselines is statistically significant for \acro{NN} Boston, despite the somewhat overlapping error bars in Figure \ref{fig:nn_boston}. 
 The improvement on \acro{NN} Cancer is more evident. 
 
\begin{table}[t]
    \centering
    \caption{Results on seven real hyperparameter tuning functions.}
    \label{table:real_function_results}
    
\begin{tabular}{lllllll}
\toprule
&{EI} & {6.EI.s} & {2-path} & {3-path} & {4-path} & {6-\eno}\\\hline
{LogReg} & 0.981   &  \textit{\textcolor{blue}{0.989  }} &  \textit{\textcolor{blue}{0.986  }} &  \textit{\textcolor{blue}{0.987  }} &  \textit{\textcolor{blue}{0.985  }} &  \textbf{0.992  } \\  
{\svm} & \textit{\textcolor{blue}{0.955  }} &  0.953   &  \textbf{0.962  } &  \textit{\textcolor{blue}{0.959  }} &  \textit{\textcolor{blue}{0.957  }} &  \textit{\textcolor{blue}{0.957  }} \\  
{\lda} & \textit{\textcolor{blue}{0.884  }} &  \textbf{0.885  } &  \textit{\textcolor{blue}{0.884  }} &  \textit{\textcolor{blue}{0.884  }} &  \textit{\textcolor{blue}{0.880  }} &  \textit{\textcolor{blue}{0.884  }} \\  
{Robot pushing 3d} & \textit{\textcolor{blue}{0.858  }} &  \textbf{0.873  } &  \textit{\textcolor{blue}{0.858  }} &  \textit{\textcolor{blue}{0.865  }} &  0.848   &  0.840   \\  
{NN Cancer} & 0.480   &  0.638   &  0.568   &  \textit{\textcolor{blue}{0.652  }} &  \textit{\textcolor{blue}{0.669  }} &  \textbf{0.683  } \\  
{NN Boston} & 0.457   &  0.461   &  \textit{\textcolor{blue}{0.475  }} &  \textit{\textcolor{blue}{0.495  }} &  \textbf{0.496  } &  \textit{\textcolor{blue}{0.485  }} \\  
{Robot pushing 4d} & \textit{\textcolor{blue}{0.408  }} &  \textit{\textcolor{blue}{0.406  }} &  \textbf{0.419  } &  \textit{\textcolor{blue}{0.413  }} &  \textit{\textcolor{blue}{0.402  }} &  0.382   \\  \hline
{Average} & 0.717   &  \textit{\textcolor{blue}{0.744  }} &  0.736   &  \textbf{0.751  } &  \textit{\textcolor{blue}{0.748  }} &  \textit{\textcolor{blue}{0.747  }} \\  
{Average (\ei<0.8)} & 0.448   &  0.501   &  0.487   &  \textit{\textcolor{blue}{0.520  }} &  \textbf{0.523  } &  \textit{\textcolor{blue}{0.517  }} \\ 
\bottomrule
\end{tabular}

\end{table}

\section{Detailed Results on Synthetic Functions}
\label{appd:syntheticfunctions}
\begin{figure}
	\centering
	\begin{subfigure}[b]{.32\textwidth}
		\includegraphics[width=\linewidth]{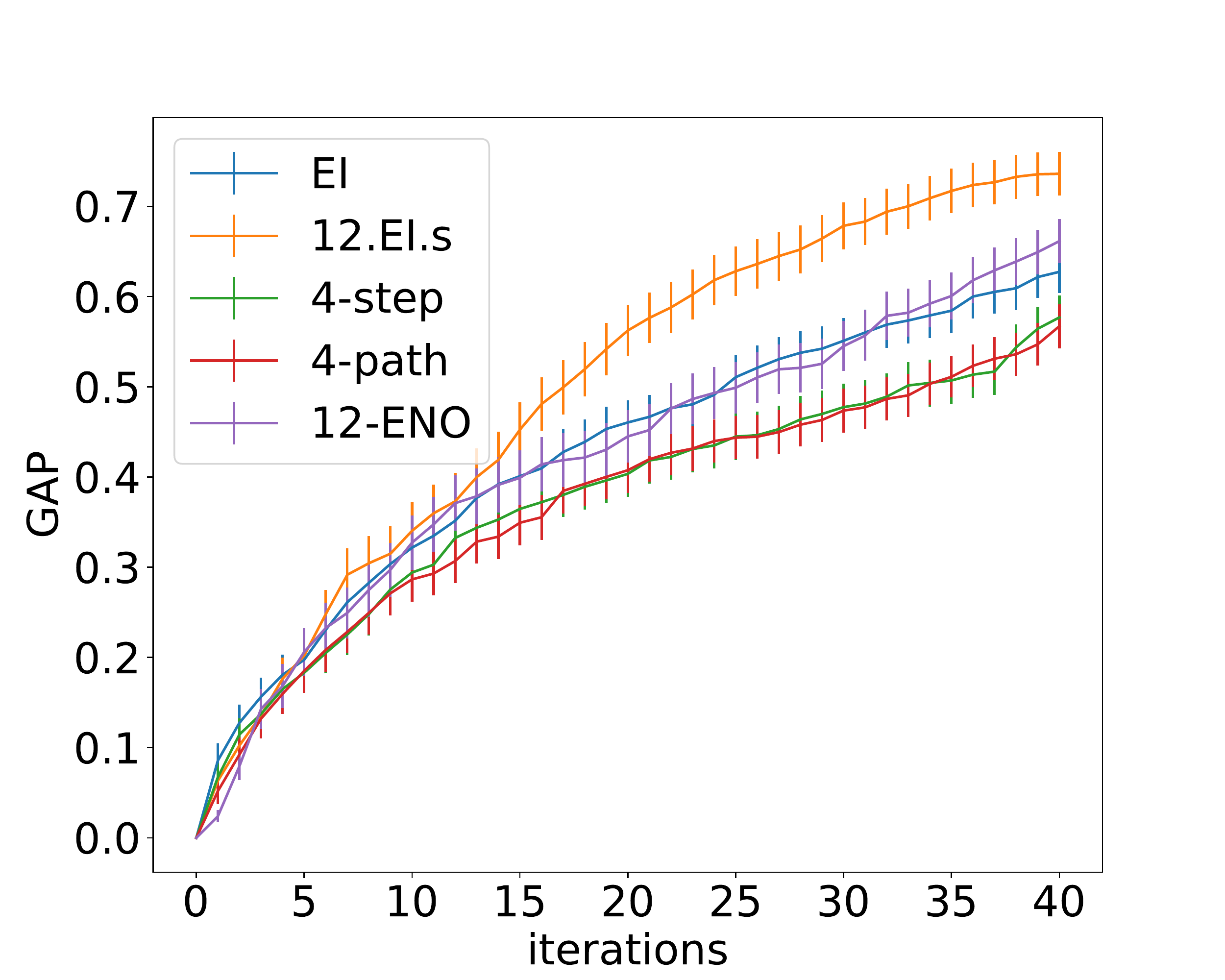}
		\subcaption{Eggholder}
		\label{fig:eggholder}
	\end{subfigure}
	\begin{subfigure}[b]{.32\textwidth}
		\includegraphics[width=\linewidth]{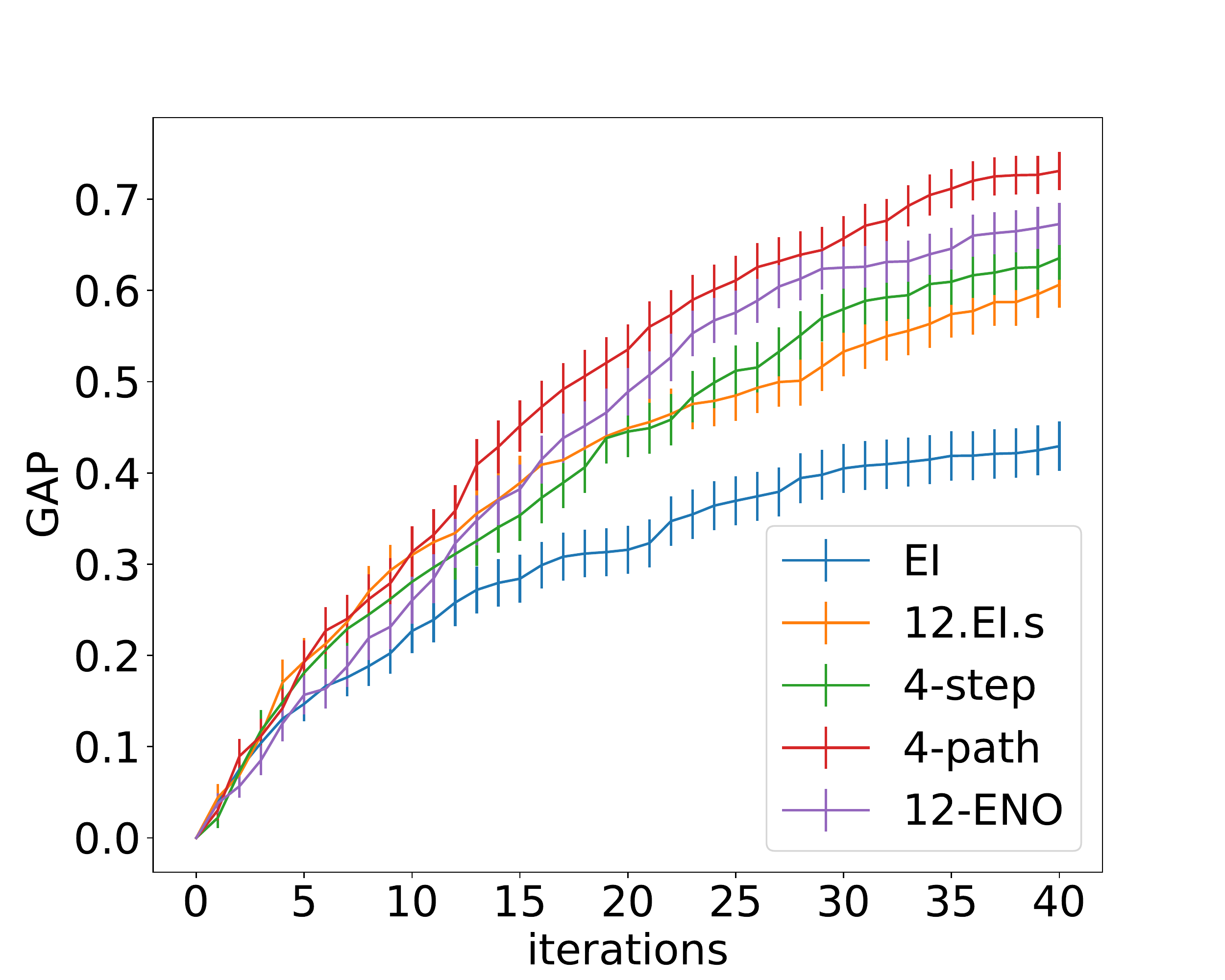}
		\subcaption{Dropwave}
		\label{fig:dropwave}
	\end{subfigure}
	\begin{subfigure}[b]{.32\textwidth}
		\includegraphics[width=\linewidth]{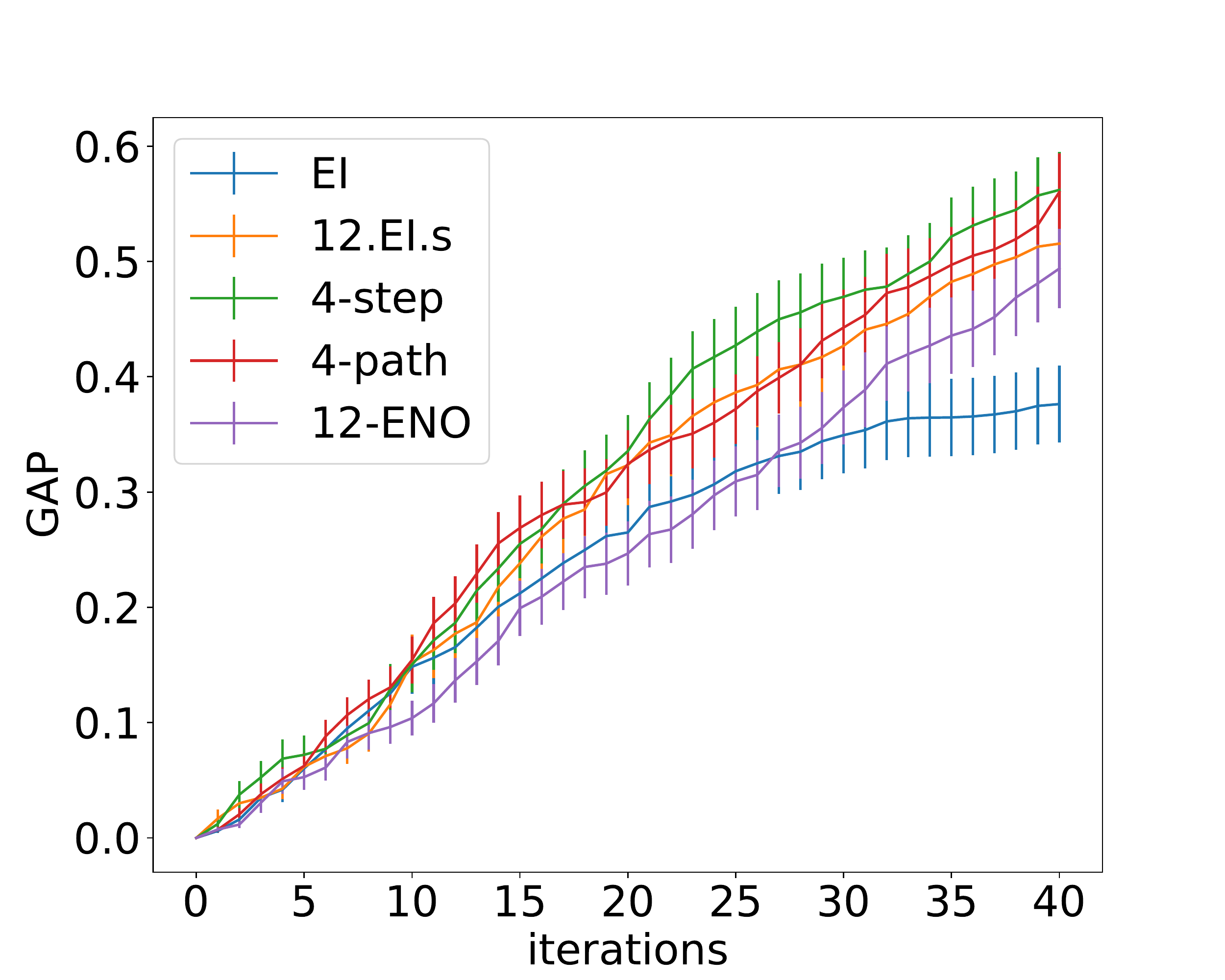}
		\subcaption{Shubert}
		\label{fig:shubert}
	\end{subfigure}
    \begin{subfigure}[b]{.32\textwidth}
		\includegraphics[width=\linewidth]{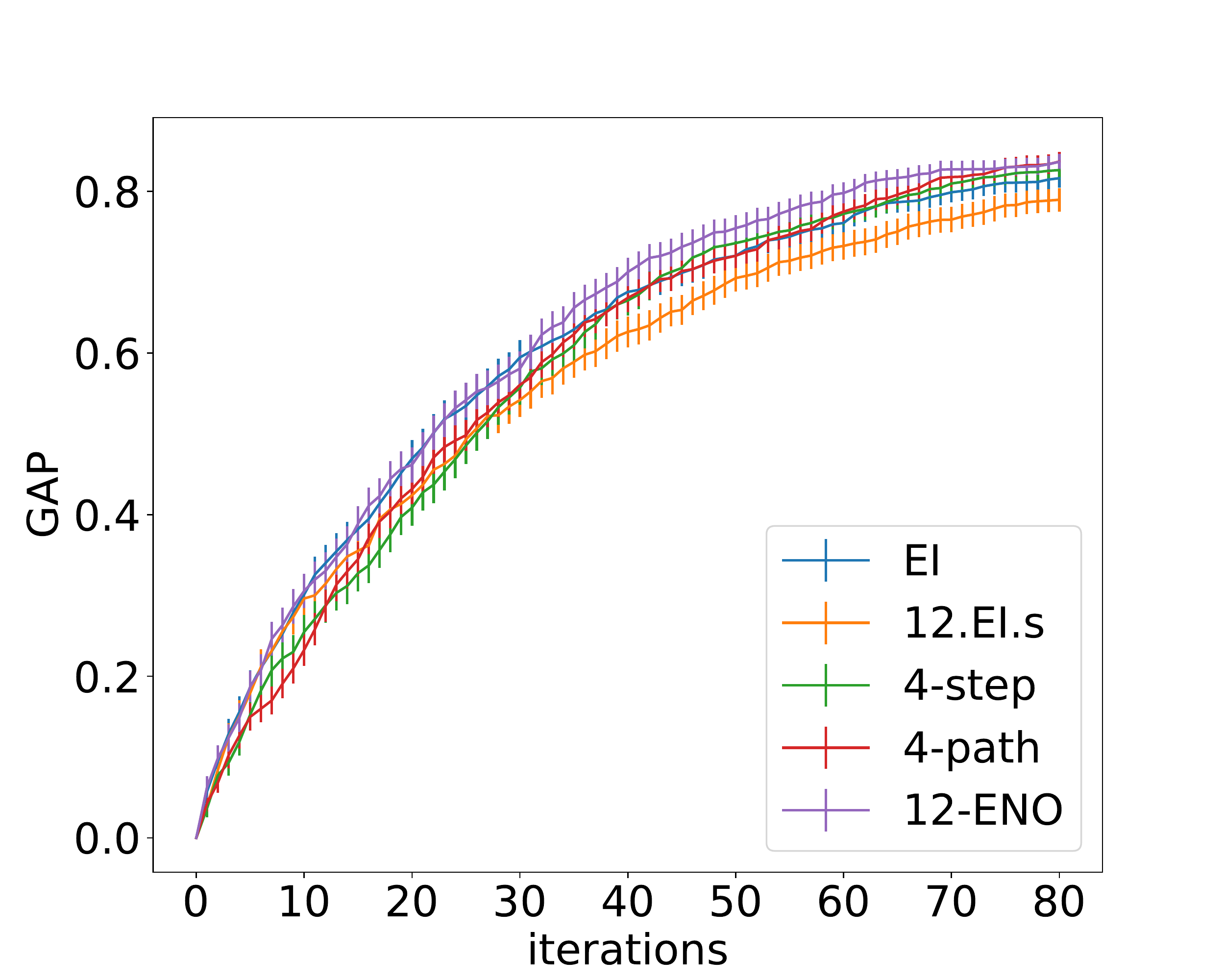}
		\subcaption{Rastrigin4}
		\label{fig:rastrigin}
	\end{subfigure}
	\begin{subfigure}[b]{.32\textwidth}
		\includegraphics[width=\linewidth]{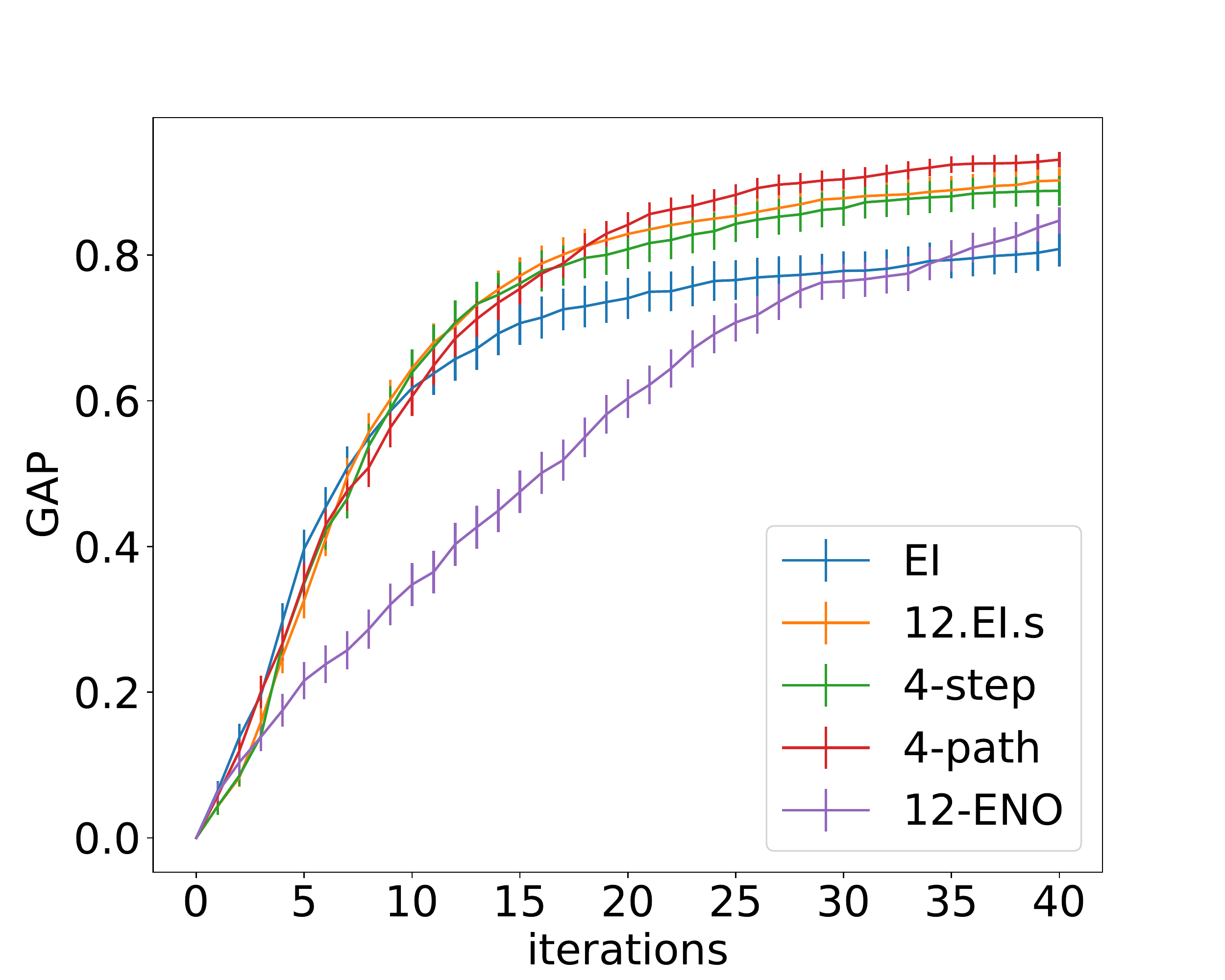}
		\subcaption{Ackley2}
		\label{fig:ackley2}
	\end{subfigure}
	\begin{subfigure}[b]{.32\textwidth}
		\includegraphics[width=\linewidth]{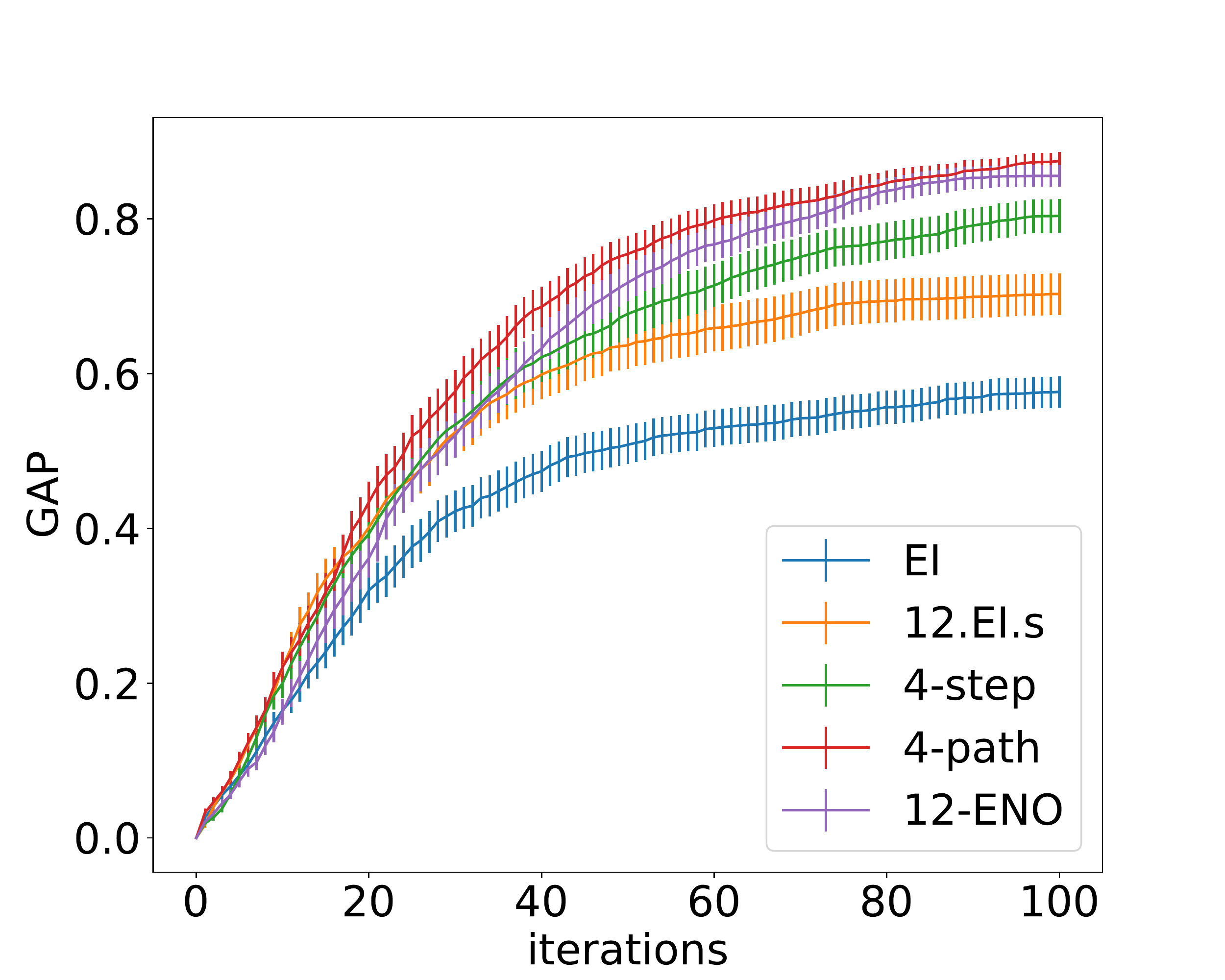}
		\subcaption{Ackley5}
		\label{fig:ackley5}
	\end{subfigure}
	\begin{subfigure}[b]{.32\textwidth}
		\includegraphics[width=\linewidth]{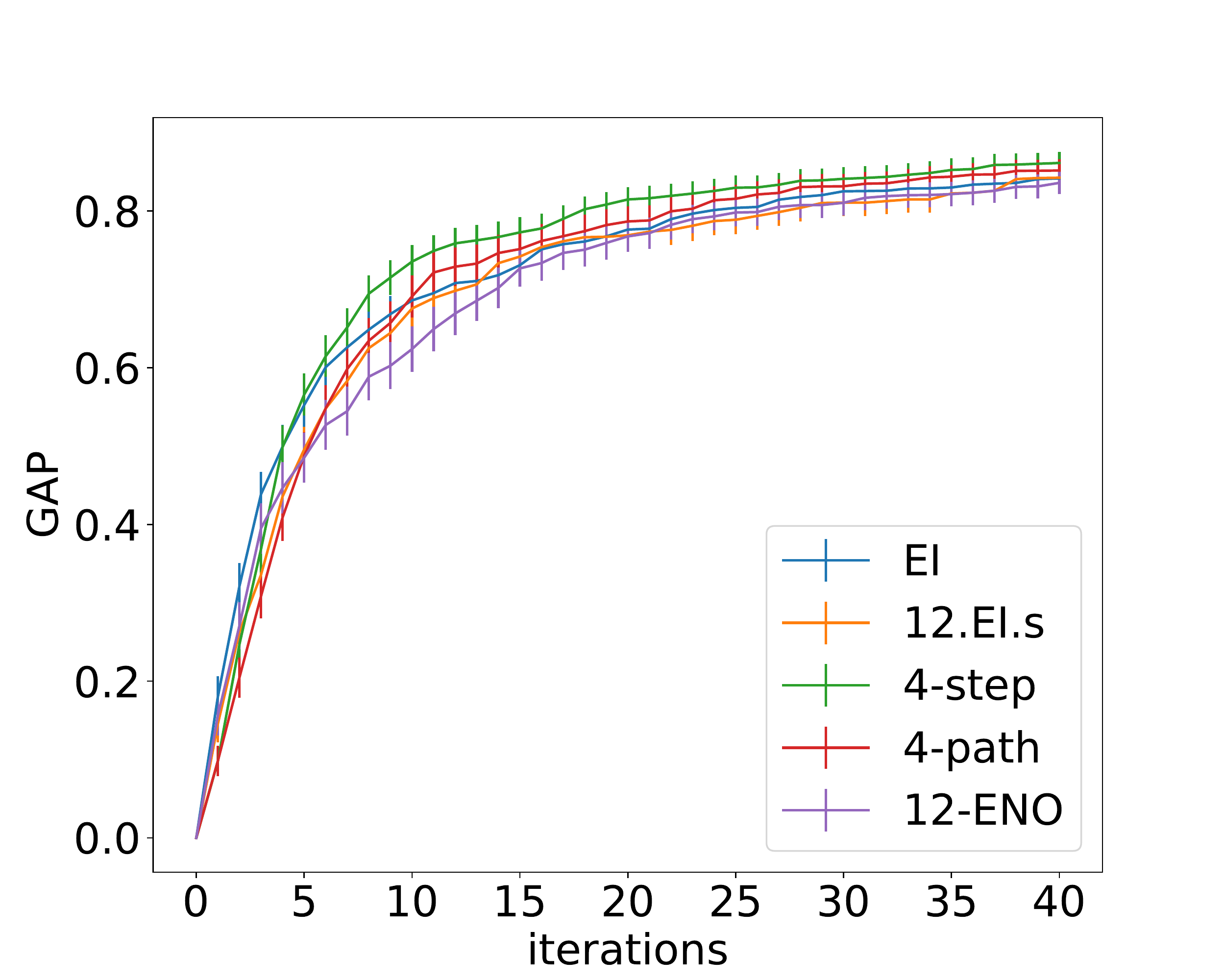}
		\subcaption{Bukin}
		\label{fig:bukin}
	\end{subfigure}
	\begin{subfigure}[b]{.32\textwidth}
		\includegraphics[width=\linewidth]{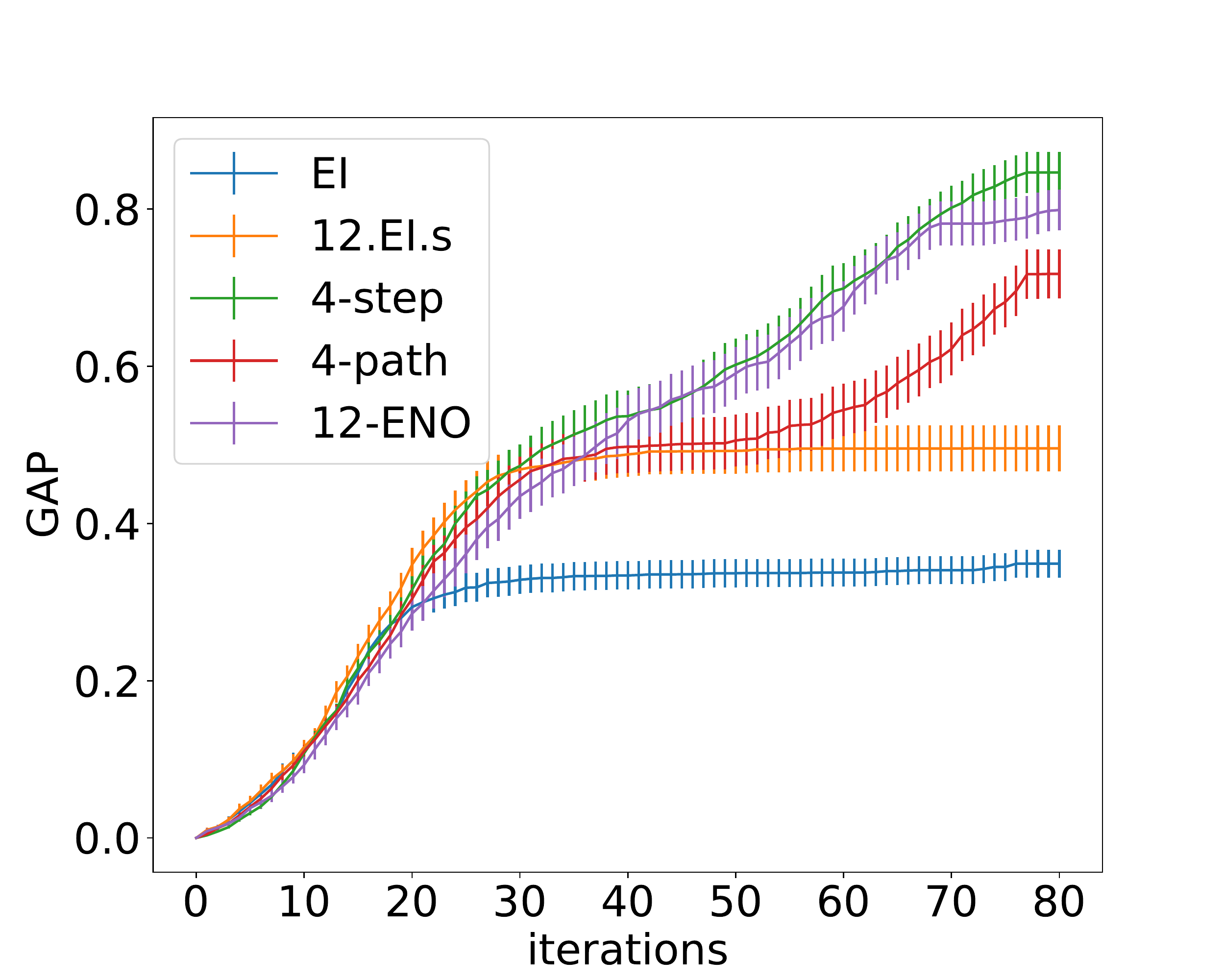}
		\subcaption{Shekel5}
		\label{fig:shekel5}
	\end{subfigure}
	\begin{subfigure}[b]{.32\textwidth}
		\includegraphics[width=\linewidth]{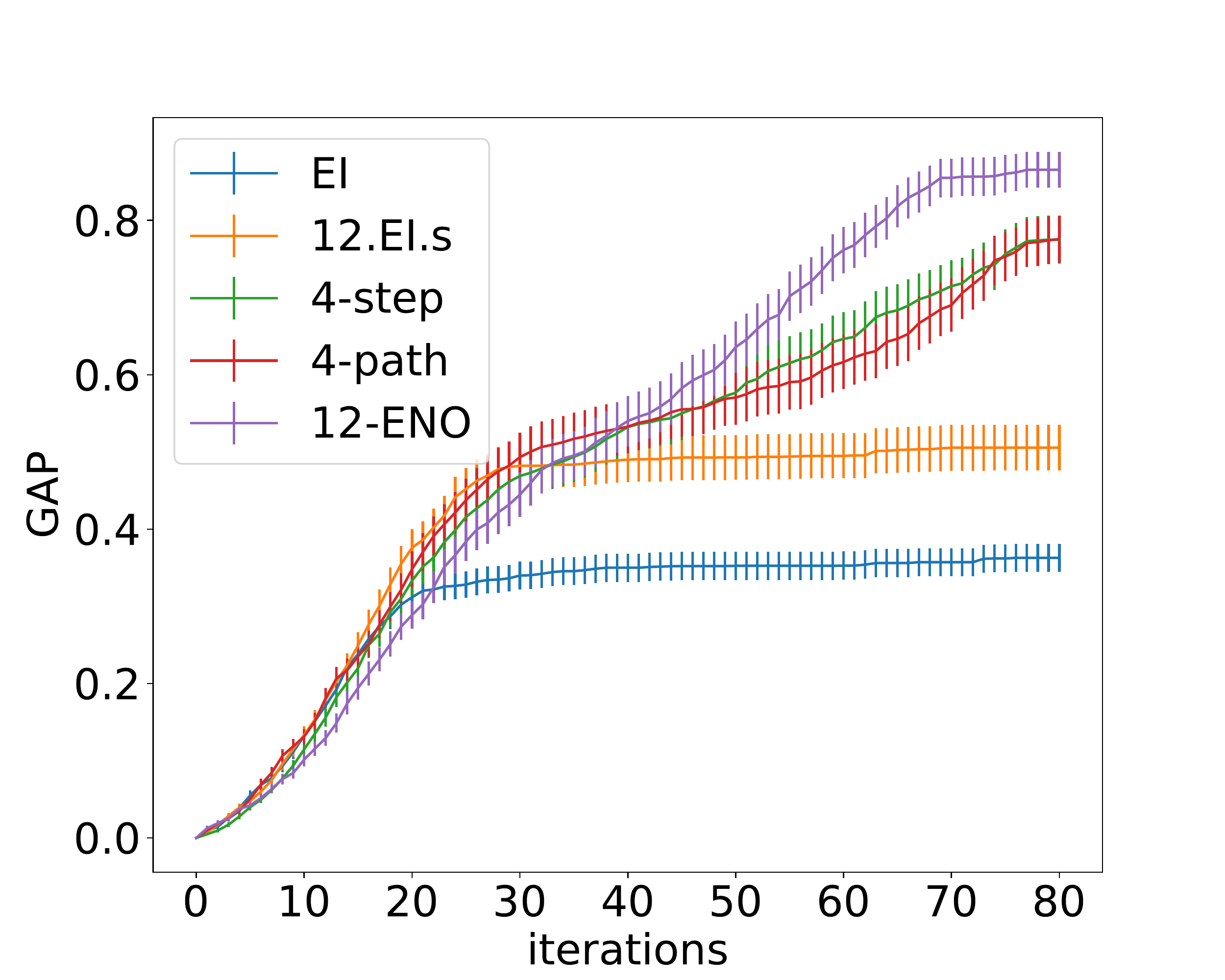}
		\subcaption{Shekel7}
		\label{fig:shekel7}
	\end{subfigure}
	\caption{
	\gap vs. iterations on individual synthetic functions.
	}
	\label{fig:gap_iter_individual_synthetic}
\end{figure}

In Figure \ref{fig:gap_iter_individual_synthetic}, we show the \gap vs. iteration plot for each individual synthetic function. 
We can see our proposed nonmyopic methods outperform baselines by a large margin on most of the functions, especially on shekel5 and shekel7.

\end{document}